\documentclass{cas-sc}

\usepackage{amsmath,amssymb,amsfonts}
\usepackage[square,numbers,sort&compress]{natbib}
\def\tsc#1{\csdef{#1}{\textsc{\lowercase{#1}}\xspace}}
\tsc{WGM}
\tsc{QE}
\tsc{EP}
\tsc{PMS}
\tsc{BEC}
\tsc{DE}

\usepackage{booktabs}
\usepackage{multirow}
\usepackage{longtable}

\usepackage{caption}
\DeclareCaptionLabelSeparator{tableNewline}{\par}
\AddToHook{env/longtable/begin}{\sffamily\selectfont}
\usepackage[dvipsnames, table]{xcolor}

\definecolor{cellbg}{HTML}{049b46}

\usepackage{graphicx} 
\usepackage{hyperref}
\hypersetup{
    hidelinks
}
\usepackage{amsthm}%
\usepackage{mathrsfs}%

\begin{document}
\let\WriteBookmarks\relax
\def\floatpagepagefraction{1}
\def\textpagefraction{.001}

\shorttitle{The Path to NDE 4.0}

\shortauthors{A. Rahman et~al.}

\title [mode = title]{The Path to NDE 4.0: A Review of Algorithms, Techniques, and Strategies for Autonomous Inspection}  

\author[1]{Ashiqur Rahman}[orcid=0000-0003-3290-2474]
\cormark[1]
\ead{ashiqur.r@niu.edu}

\credit{Conceptualization of this research, data curation, literature review, and manuscript preparation.}

\author[1]{Venkata Devesh Reddy Seethi}[orcid=0000-0002-7400-7518]
\ead{dseethi@niu.edu}
\credit{Data curation and manuscript preparation.}

\author[2]{Zachary Kral}[orcid=0000-0003-4188-4036]
\ead{zachary.t.kral@spiritaero.com}
\credit{Data curation and manuscript preparation.}

\author[3]{Austin Yunker}[orcid=0009-0004-7253-6370]
\ead{ayunker@anl.gov}
\credit{Data curation.}

\author[3]{Rajkumar Kettimuthu}[orcid=0000-0002-0046-9883]
\ead{kettimut@mcs.anl.gov}
\credit{Co-ordination and research supervision.}

\author[1]{Hamed Alhoori}[orcid=0000-0002-4733-6586]
\ead{alhoori@niu.edu}
\credit{Co-ordination, research supervision, and manuscript review.}

\affiliation[1]{organization={Northern Illinois University},
                city={DeKalb},
                postcode={60115}, 
                state={IL},
                country={USA}}

\affiliation[2]{organization={Spirit AeroSystems (Boeing)},
                city={Wichita},
                postcode={67210}, 
                state={KS},
                country={USA}}

\affiliation[3]{organization={Argonne National Laboratory},
                city={Lemont},
                postcode={60439}, 
                state={IL},
                country={USA}}

\cortext[cor1]{Corresponding author.}

\begin{abstract}
    Non-Destructive Inspection (NDI) is a cornerstone of modern quality assurance, critical for ensuring the safety and reliability of infrastructure and manufactured components without causing damage. This paper presents a comprehensive review of the current NDI landscape, analyzing the field through four distinct lenses, namely, computational algorithms, physical techniques, industrial applications, and research objectives. We synthesize a broad range of academic literature to illustrate the rapid evolution from manual, analog interpretation to a digitized, data-centric paradigm known as NDE 4.0. Our analysis highlights the dominance of ultrasound and visual inspection as primary physical techniques, while revealing a significant shift toward deep learning and generative models for automated defect recognition. Despite these technological strides, the industry faces critical bottlenecks, notably the scarcity of standardized open-access datasets and the ``black box'' nature of AI models, which hinders widespread adoption. By charting these developments and identifying key research gaps, such as the need for explainable AI (XAI) and standardized digital twin protocols, this review offers a strategic roadmap for future research. It concludes that the future of NDI research should focus on the successful integration of intelligent, autonomous fusion of inspection data with advanced sensor technologies into asset lifecycle management.
\end{abstract}






\maketitle



\section{Introduction} \label{sec:review-introduction}

\begin{figure}[pos=htbp]
    \centering
    \includegraphics[width=\textwidth]{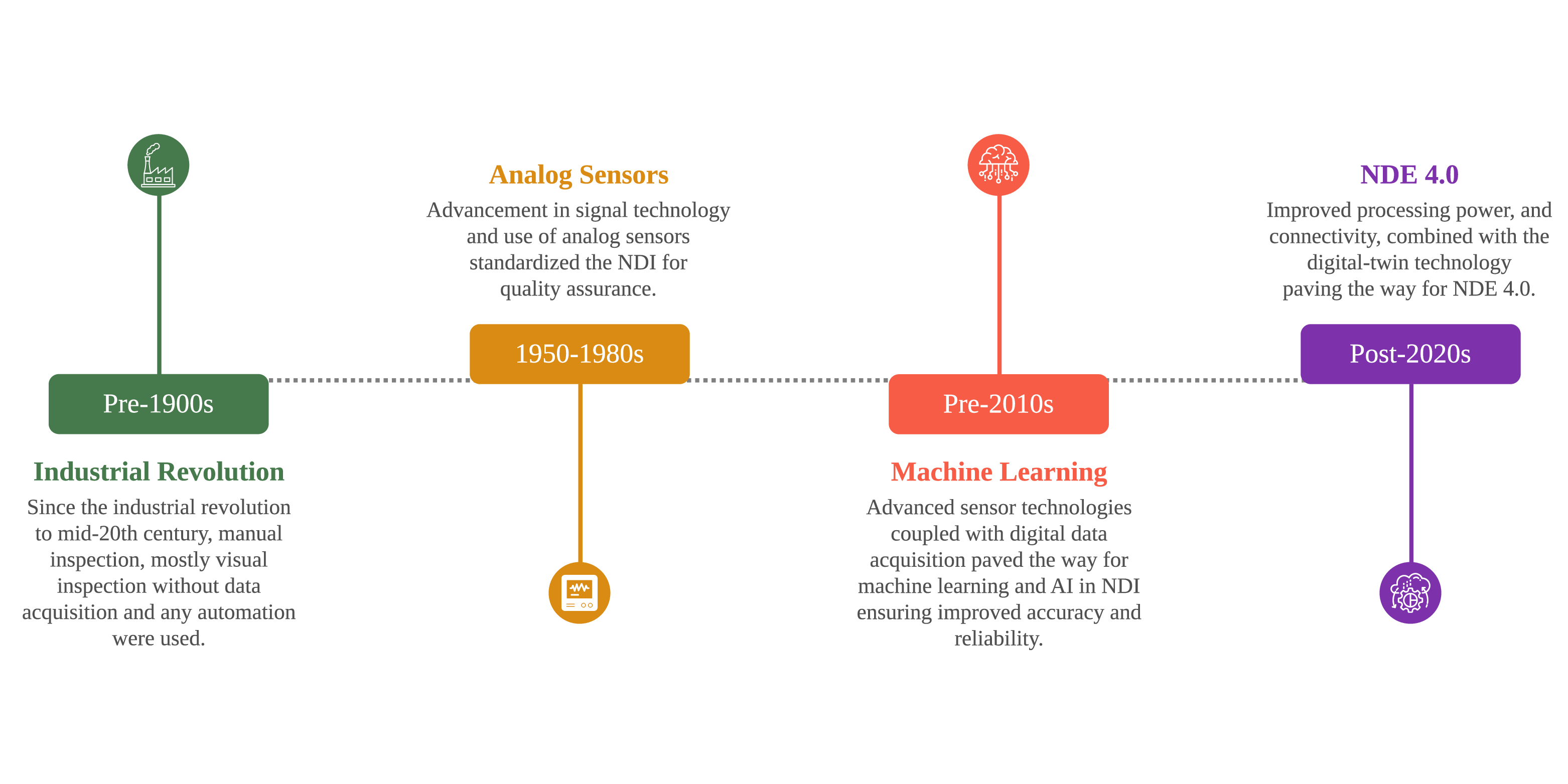}
    \caption{Progress of NDI over time.}
    \label{fig:review-progress-of-ndi}
\end{figure}

The process of non-destructive inspection (NDI) is an integral component of modern manufacturing and quality assurance, though its origins extend back for centuries. While its simplest form, direct visual inspection, is a practice as old as manufacturing itself, the Industrial Revolution and the technological advancements that followed spurred the rapid evolution and automation of NDI techniques. Consequently, a vast body of research has emerged to standardize, improve, and perfect NDI processes across diverse industries \citep{Rummel2012-zl, Chauveau2018-qo, Honarvar2020-tw, Dwivedi2018-ts, Gholizadeh2016-ww}. This field is currently undergoing a fundamental transformation. Historically characterized by manual data interpretation and analog measurement techniques \citep{Aastroem2008-ew, Maldague1993-wx}, the discipline is rapidly migrating toward a digitized, automated, and data-centric paradigm often termed \textit{NDE 4.0} \citep{Meyendorf2023-qv}. This shift is driven by the convergence of advanced sensor physics, high-performance computing, and artificial intelligence. Recent academic literatures reflect this transition intensely, revealing a landscape where the boundaries between physical inspection and computational modeling are becoming increasingly blurred.

The primary objective of modern NDI research is no longer merely the detection of macroscopic flaws. Rather, it is the quantitative characterization of material states, the prediction of remaining useful life (RUL), and the integration of inspection data into the broader digital twin ecosystem of manufacturing and infrastructure management \citep{Meyendorf2023-qv}. Figure \ref{fig:review-progress-of-ndi} shows the progress of NDI over time and the future direction of the field. 

This study provides a comprehensive review of the available scientific literature on NDI, charting the landscape of current techniques and identifying the various industries that depend on these methods. To organize this analysis effectively, the literature is divided into four distinct categories, namely, Algorithms, Techniques, Application Areas, and Research Type. The \textit{Algorithms} category investigates computational methods designed to automate the NDI process. The \textit{Techniques} category encompasses literature regarding the physical technologies and methodologies applied during inspection. Papers within the \textit{Application Areas} category examine NDI applications and the current technological standards across various industrial sectors. Finally, the \textit{Research Type} category groups studies according to their primary research objectives. This structured classification highlights both the established strengths and the ongoing challenges within NDI research. The analysis indicates that the most significant bottleneck in modern NDI is no longer sensor sensitivity, but data interpretation and the chronic scarcity of defective training data in industrial settings. This analysis is intended to serve as a valuable resource, offering a structured perspective on the current state-of-the-art in non-destructive inspection, that can help future researchers identify critical problems to solve and, in doing so, advance the industry forward.

\section{Background}
This study synthesizes impactful academic research from diverse disciplines that collectively shape the modern NDI landscape. To provide a clear and structured overview of this broad body of work, Table \ref{tab:review-overview} summarizes the selected literature, grouping each paper into its primary research area and contribution.

\begin{table}[pos=htbp]
    \centering
    \caption{Overview of the reviewed papers.}
    \begin{tabular}{llr}
    \toprule
        Research Area & Category & No. of Papers \\
    \midrule
        \multirow{4}{4em}{Algorithms} & Computer Vision and Image Processing & 16 \\
         & Machine Learning & 8 \\
         & Deep Learning & 10 \\
         & Anomaly Detection & 4 \\
         & Generative Network & 4 \\
     \midrule
        \multirow{6}{4em}{NDI Techniques} & Ultrasound & 22 \\
        & Radiography and X-Ray & 11 \\
        & Infrared, Thermal, and Low Frequency & 15 \\
        & Visual Inspection & 7 \\
        & Electromagnetic Field & 8 \\
    \midrule
        \multirow{6}{4em}{Application Areas} & Composite Materials & 10 \\
        & Organic Material and Food Safety & 4 \\
        & Construction Material & 2 \\
        & Fragile Artifacts & 4 \\
        & Medical & 2 \\
        & Structural Health Monitoring & 11 \\
    \midrule
        \multirow{3}{4em}{Literature Type} & Review & 21 \\
        & Proposal of New Method & 13 \\
        & Evaluating Methods & 12 \\
    \bottomrule
    \end{tabular}
    \label{tab:review-overview}
\end{table}

\subsection{Definitions} \label{sec:review-definitions}
To ensure clarity and precision throughout this review, we first establish consistent definitions for the key technical terminologies and concepts central to the field of NDI. This section defines these fundamental terms and clarifies their specific usage within the context of this paper.

\subsubsection*{Non-destructive Inspections}
Non-Destructive Inspection (NDI) defines the process of evaluating a material, component, or system without inducing damage or altering its future usefulness. Techniques in this domain range from simple surface examinations to complex volumetric analyses capable of revealing deep subsurface defects \citep{Dwivedi2018-ts}. While the terms Non-Destructive Testing (NDT) and Non-Destructive Evaluation (NDE) are often used interchangeably in the literature, this paper consistently employs \textit{NDI}, reserving the use of \textit{NDE} exclusively for discussions regarding ``NDE 4.0.''

\subsubsection*{Computer Vision}
Computer Vision is an interdisciplinary scientific field that focuses on enabling computers to gain a high-level understanding from digital images or videos. By employing computational models to simulate human visual perception, this technology automates the extraction, analysis, and processing of useful information from visual data \citep{Voulodimos2018-zr}. In the context of NDI, computer vision transforms subjective visual inspections into objective, quantitative analyses, allowing for the automated detection and segmentation of defects in both optical images and visualized sensor data.

\subsubsection*{Deep Learning}
Deep learning (DL) is a machine learning method that uses computational models with multiple processing layers to learn representations of data at multiple levels of abstraction. Each layer learns and combines complex features from a large dataset to achieve an assigned goal \citep{LeCun2015-kr}.

\subsubsection*{Machine Learning}
Machine learning (ML) is a subfield of artificial intelligence (AI) where the computer algorithm learns from given data to perform specific tasks. The algorithm can identify patterns, make predictions, or extrapolate new information without explicit programming for each situation \citep{Jordan2015-nw}.

\subsubsection*{POD}
In non-destructive inspection (NDI), Probability of Detection (POD) is used to quantify the effectiveness of a technique, which refers to the statistical likelihood of detecting a flaw or defect of a certain size and type using a specific NDI method \citep{Rummel2012-zl}.

\subsubsection*{Structural Health Monitoring}
Structural Health Monitoring (SHM) is a system designed to prevent the failure of safety-critical structures, such as aircraft, bridges, and nuclear reactors. It uses NDI techniques and a large number of sensors to provide continuous or on-demand information about the state of a structure. This allows for an assessment of the structural integrity at any time so that timely remedial actions can be taken as needed \citep{Achenbach2009-jj}.

\subsubsection*{Asset Lifecycle Management}
Asset Lifecycle Management (ALM) is a strategic framework for managing the entire lifespan of a physical asset, from design and manufacturing through to decommissioning, to maximize its operational value and safety. In the context of NDE 4.0, ALM serves as the organizational goal, where digitized inspection data is integrated into digital twins to enable predictive maintenance and informed decision-making throughout the asset's life \citep{Meyendorf2023-qv}.

\subsubsection*{NDE 4.0}
NDE 4.0 is the convergence of Non-Destructive Evaluation (NDE) with the digital technologies of Industry 4.0. It represents a shift from analog, manual inspections to a digitized, interconnected ecosystem where inspection data is automatically collected, analyzed via AI/ML, and integrated into ``Digital Twins" to enable predictive maintenance and data-driven decision-making throughout a product's lifecycle \citep{Meyendorf2023-qv}.

\subsubsection*{Composite Materials}
A composite material is a material made by combining two or more different substances that, when joined, create a new material with superior characteristics, such as being stronger, lighter, or more durable than the individual components. The original materials do not dissolve or blend into each other. Instead, one material surrounds and holds in place the other, allowing them to work together to enhance performance \citep{Gholizadeh2016-ww}.

\subsubsection*{Ultrasound}
Ultrasound in non-destructive inspection refers to high-frequency sound waves, used in Ultrasonic Testing (UT) to characterize materials and identify internal defects for quality control \citep{Hassani2023-hi}. This method works by emitting ultrasonic impulses, often between $0.2$ and $15 MHz$, from a transducer into an object. By analyzing the received signals and detecting changes caused by reflection, attenuation, or energy dissipation, inspectors can deduce the presence of internal flaws, measure material thickness, or determine other properties without damaging the component.

To effectively visualize these internal characteristics, ultrasonic data is typically presented in three formats, A-scan, B-scan, and C-scan. An \textbf{A-scan} is a one-dimensional representation of the signal amplitude versus time, showing the depth of indications at a single point. A \textbf{B-scan} collects multiple A-scans along a line to form a two-dimensional cross-sectional view of the material. Finally, a \textbf{C-scan} maps signal features, such as amplitude or time-of-flight, across a scanned surface area to create a two-dimensional planar image, providing a top-down view of defects within the component \citep{Yunker2025-ck}.

\subsubsection*{Piezoelectric Sensor}
A piezoelectric sensor is a device that uses the piezoelectric effect to measure changes in pressure, acceleration, strain, or force by converting them into an electrical charge. This effect occurs in certain materials, which generate a voltage when mechanical stress is applied to them \citep{Niezrecki2014-of}.

\subsubsection*{Impedance Method}
The impedance method is a non-destructive testing technique used to find hidden flaws like cracks or delaminations inside a material. It works by tapping the material at a single point with a steady, low-frequency vibration and simultaneously measuring how much the material resists that vibration \citep{Cakir2025-fa}.

\section{Algorithms in NDI} \label{sec:review-algorithms-in-ndi}

\begin{figure}[pos=htp]
    \centering
    \includegraphics[width=1\linewidth]{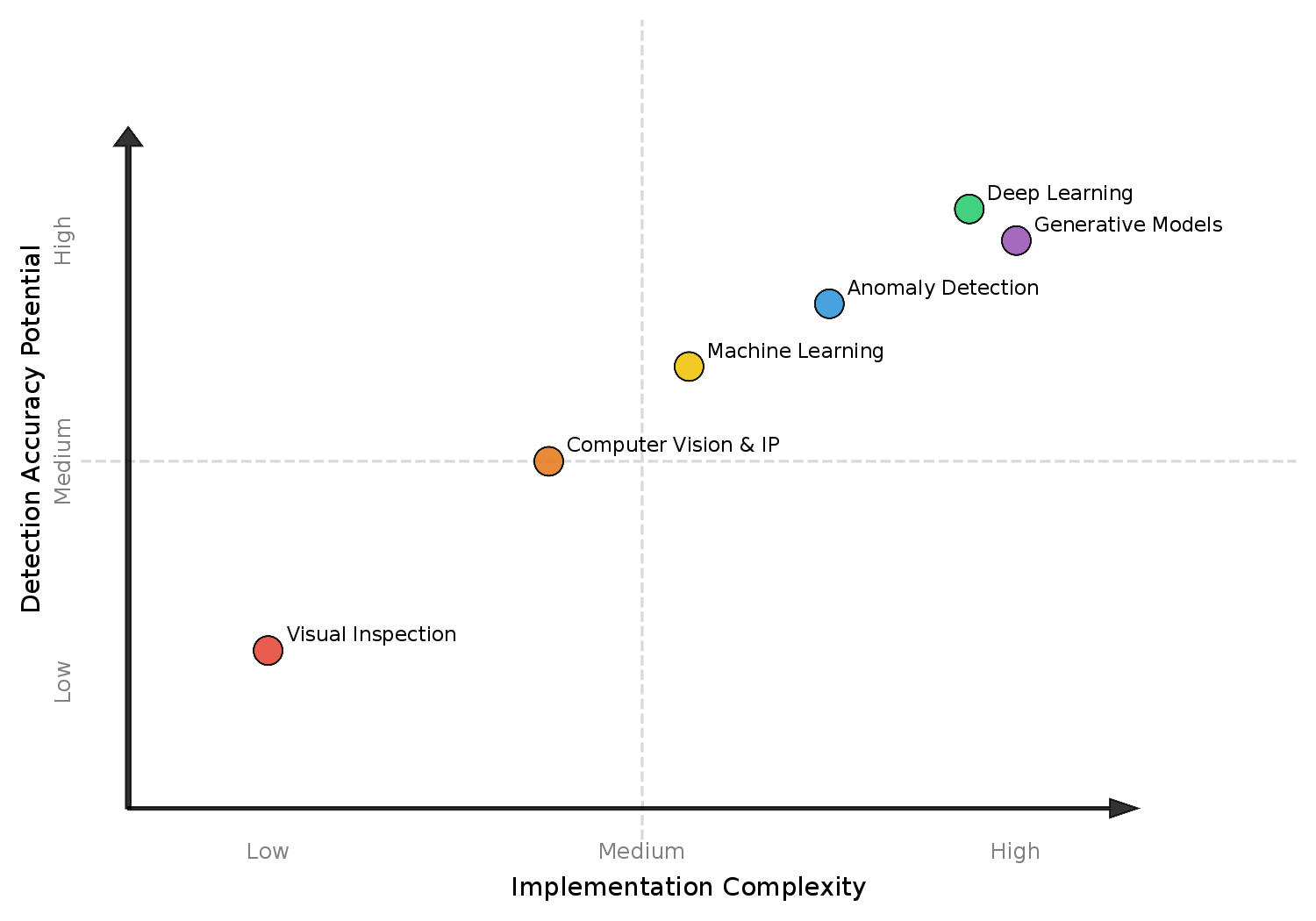}
    \caption{Qualitative comparison of NDI algorithms.}
    \label{fig:review-ndi-algorithm-qualitative-comparison}
\end{figure}

The advancement of computational intelligence has dawned a paradigm shift in NDI, moving the field from a reliance on manual interpretation to an era of automated, data-driven analysis. This transition is driven by the need to process the vast and complex datasets generated by modern NDI techniques with greater speed, objectivity, and precision. In response, a significant body of research has focused on developing and applying sophisticated algorithms to enhance every stage of the inspection process. This section provides a comprehensive overview of these algorithmic advancements, charting the evolution from foundational computer vision and image processing techniques to the more complex and powerful paradigms of ML, DL, and generative modeling. Table \ref{tab:review-algorithm-paper-domain} provides a categorical breakdown of the surveyed literature, mapping each study to its primary focus, ranging from foundational computer vision techniques to advanced generative modeling strategies. Figure \ref{fig:review-ndi-algorithm-qualitative-comparison} illustrates a qualitative comparison of the algorithmic methods based on the discussion in this section.

\begin{longtable}[c]{lp{0.8cm}p{0.8cm}p{0.8cm}p{0.8cm}p{0.8cm}}
    \caption{Domains of the research papers discussed in Section \ref{sec:review-algorithms-in-ndi}. Domain labels are CVIP = Computer Vision and Image Processing, ML = Machine Learning, DL = Deep Learning, AD = Anomaly Detection, and GM = Generative Models. \label{tab:review-algorithm-paper-domain}} \\
    \toprule
         Paper & CVIP & ML & DL & AD & GM \\
    \midrule
    \endfirsthead
    \toprule
         Paper & CVIP & ML & DL & AD & GM \\
    \midrule
    \endhead
         \citet{Sezgin2001-cc} & \cellcolor{cellbg} & & & & \\
         \citet{Huang2003-gy} & \cellcolor{cellbg} & & & & \\
         \citet{Brosnan2004-yd} & \cellcolor{cellbg} & & & & \\
         \citet{Dolinko2008-ds} & \cellcolor{cellbg} & & & & \\
         \citet{Memmolo2015-er} & \cellcolor{cellbg} & & & & \\
         \citet{Yousefi2018-el} & & & \cellcolor{cellbg} & & \\
         \citet{Akcay2019-xy} & \cellcolor{cellbg} & & \cellcolor{cellbg} & \cellcolor{cellbg} & \cellcolor{cellbg} \\
         \citet{Sreeshan2019-sf} & \cellcolor{cellbg} & & & & \\
         \citet{Baumgartl2020-fe} & & & & \cellcolor{cellbg} & \\
         \citet{Gardner2020-bn} & & \cellcolor{cellbg} & & & \\
         \citet{Ma2020-cu} & & & \cellcolor{cellbg} & & \\
         \citet{Mishra2020-qv} & & \cellcolor{cellbg} & & & \\
         \citet{Siegel2020-cp} & & & \cellcolor{cellbg} & & \\
         \citet{Sreeshan2020-dc} & \cellcolor{cellbg} & & & & \\
         \citet{Tabernik2020-mr} & \cellcolor{cellbg} & & & & \\
         \citet{Zhang2020-kp} & \cellcolor{cellbg} & & & & \\
         \citet{Alloqmani2021-gz} & & & & \cellcolor{cellbg} & \\
         \citet{Fang2021-nw} & & & & & \cellcolor{cellbg} \\
         \citet{Hashimoto2021-mf} & \cellcolor{cellbg} & & & & \cellcolor{cellbg} \\
         \citet{Na2021-eq} & & & \cellcolor{cellbg} & & \\
         \citet{Mahanti2022-sp} & \cellcolor{cellbg} & & & & \\
         \citet{Posilovic2022-ay} & & & \cellcolor{cellbg} & & \\
         \citet{Presenti2022-kp} & & & \cellcolor{cellbg} & \cellcolor{cellbg} & \\
         \citet{Taheri2022-ix} & & \cellcolor{cellbg} & & & \\
         \citet{Yang2022-nl} & & \cellcolor{cellbg} & & &  \\
         \citet{Ali2023-cp} & \cellcolor{cellbg} & & & &  \\
         \citet{Angermann2023-at} & & & & & \cellcolor{cellbg} \\
         \citet{Hola2023-sh} & & \cellcolor{cellbg} & & &  \\
         \citet{Lin2023-fd} & & \cellcolor{cellbg} & & &  \\
         \citet{Sun2023-yp} & & \cellcolor{cellbg} & \cellcolor{cellbg} & &  \\
         \citet{Tabernik2023-dp} & \cellcolor{cellbg} & & \cellcolor{cellbg} & & \\
         \citet{Barshok2025-be} & & & \cellcolor{cellbg} & & \\
         \citet{Rahman2025-wf} & \cellcolor{cellbg} & & & & \\
         \citet{Ribeiro2025-mf} & \cellcolor{cellbg} & & & & \\
         \citet{Yang2025-rz} & & \cellcolor{cellbg} & & & \\
         
    \bottomrule
\end{longtable}

\subsection{Computer Vision and Image Processing} \label{sec:review-computer-vision-and-image-processing}

While visual inspection is one of the oldest forms of quality control, the integration of modern computer vision and image processing has revolutionized the field, enabling automated, objective, and highly sensitive non-destructive inspection. The core objective of these techniques is to computationally analyze image-based data from various NDI modalities to detect, segment, and classify defects. This process typically involves a pipeline of operations, from initial image enhancement and visualization to sophisticated, algorithm-driven defect recognition.

A critical first step in any automated image-based NDI is isolating potential defects from the background, a task known as image segmentation. In a foundational study, \citet{Sezgin2001-cc} surveyed numerous image thresholding methods, which are fundamental to segmentation in NDT applications. By evaluating thresholding techniques based on misclassification error, relative foreground area error, and signal-to-noise ratio, they identified that a mixture method for thresholding offers generalizable and robust performance for accurately identifying defective regions.

Beyond segmentation, enhancing the raw NDI data is often necessary to improve the visibility of subtle flaws. Research has explored various enhancement techniques tailored to specific NDI modalities. For instance, \citet{Sreeshan2019-sf} proposed a method combining Gabor filters and watershed segmentation to increase the contrast of debonding defects in thermographic images of composite laminates. In a subsequent study \citep{Sreeshan2020-dc}, they developed a multi-stage process for pulsed thermography that included pre-processing for image enhancement, followed by defect candidate identification and precise segmentation, successfully identifying $14$ out of $16$ simulated defects in an aerospace composite. Similarly, \citet{Dolinko2008-ds} presented an image processing approach to improve the visualization of defect borders in flawed plates by analyzing temperature distribution under a thermal load. For Digital Speckle Pattern Interferometry (DSPI). \citet{Memmolo2015-er} developed an automated process to reduce distortions from external noise by smoothing interferograms with a Gaussian filter and analyzing their Fourier spectrum, a method particularly useful for monitoring fragile artifacts like wooden paintings.

Additionally, many advanced NDI techniques, such as industrial computed tomography (CT) and ultrasonic B-scans, produce complex, high-dimensional data that requires sophisticated visualization methods for human interpretation and algorithmic analysis. \citet{Huang2003-gy} addressed the challenge of visualizing noisy, high-resolution volume data from industrial CT. They proposed a novel interface combining region-growing techniques with a 2D histogram to facilitate volumetric feature extraction, enabling users to identify, separate, and model concealed structures within objects. Similarly, \citet{Zhang2020-kp} tackled the difficulty of identifying defects in ultrasonic B-scan images of composite materials where fiber waviness can reduce defect reflection by up to 50\%. By simulating and visualizing signals from a fabricated composite sample, they proposed that combining amplitude data with B-scan images leads to more reliable defect detection in such complex materials. Texture analysis has also emerged as a sophisticated sub-field of image processing, in the context of corrosion detection. Unlike cracks, which often have distinct edges, corrosion presents as an irregular, texture-based anomaly. Recent papers discuss the development of texture-energy measures and filtering approaches designed to identify the edgeless boundaries of corroded regions on metallic surfaces \citep{Ali2023-cp}. This represents a significant maturation of CV techniques, moving beyond simple color thresholding to morphological analysis that can quantify the severity of material degradation.

The primary goal of applying computer vision to NDI is to automate the inspection process, bringing the benefits of speed, consistency, and objectivity that surpass human capabilities. The food industry provides a clear example of this, where computer vision is used for applications ranging from defect detection and quality grading to sorting and species identification. Reviews by \citet{Brosnan2004-yd} and \citet{Mahanti2022-sp} highlight the success of NDI in food and fruit inspection, emphasizing its ability to provide real-time damage assessment for bruises, cracks, and internal defects without harming the product. A key takeaway from this research is the critical importance of a well-controlled image acquisition environment, including proper illumination, for the success of any computer vision-based system.

More recently, research has shifted toward advanced deep learning architectures, particularly generative models, to address two of the most significant challenges in NDI, namely, the scarcity of labeled defect data and the detection of novel or unexpected anomalies. For anomaly detection, \citet{Akcay2019-xy} introduced GANomaly, a semi-supervised method based on Generative Adversarial Networks (GANs). By training a GAN on images of non-defective components, the model learns the distribution of normal data. When presented with an image containing a defect, the model fails to reconstruct it accurately, signaling an anomaly. This approach is highly effective as it requires only a limited amount of labeled data, or none at all, to achieve high accuracy. \citet{Hashimoto2021-mf} also employed a deep generative model to identify cracks in magneto-optical (MO) images of steel specimens, successfully identifying slit defects with a width of just $0.15mm$ and a depth of $6mm$. The field has been further advanced by specialized deep learning architectures, like segmentation-based models that automatically learn to identify defect features. These models outperform traditional methods and facilitate automated quality control \citep{Tabernik2020-mr, Tabernik2023-dp}. Recent studies have demonstrated the feasibility of using instance segmentation architectures, such as Mask R-CNN and YOLO variants, to parse complex ultrasonic scans of aerospace composites \citep{Rahman2025-wf}. \citet{Ribeiro2025-mf} also demonstrated the use of automated visual defect detection in the context of pharmaceutical bottles production. These methods represent a significant leap forward, moving from engineered feature detection to automated feature learning for more robust and accurate NDI.

\subsection{Machine Learning} \label{sec:review-machine-learning}
The adoption of ML has transformed non-destructive inspection, advancing capabilities from basic automated analysis to complex predictive modeling and data-driven decision-making that significantly enhance the efficiency and reliability of the NDI process. Recent academic literature reflects a dual focus in this area. On one hand, studies have rigorously evaluated and benchmarked the performance of specific ML models for NDI tasks \citep{Mishra2020-qv}. Simultaneously, a significant body of research has been dedicated to broader discussions on the strategic importance, practical utility, and inherent challenges of integrating these advanced analytical techniques across various NDI applications \citep{Gardner2020-bn, Taheri2022-ix, Lin2023-fd}. Despite the rapid rise of DL, classical ML algorithms continue to maintain a strong presence in the literature due to their robustness and efficiency. \citet{Hola2023-sh} established that the random forest algorithm provides the best performance for predicting the moisture content of concrete and masonry structures. This conclusion is supported by recent work from \citet{Yang2025-rz} who evaluated multiple algorithms on two different datasets and found that Random Forest outperformed more complex models.

A primary application of machine learning in NDI involves predicting material properties based on sensor data. For example, a study by \citet{Mishra2020-qv} assessed the performance of three models, specifically artificial neural networks (ANN), adaptive neuro fuzzy inference systems (ANFIS), and support vector regression (SVR), for predicting the compressive strength of masonry in historic buildings. Their research, which used data from ultrasonic pulse velocity and rebound hammer tests, concluded that the SVR model delivered the most accurate predictions, presenting a viable method for structural evaluation. Machine learning is also widely used for classification tasks, such as identifying subsurface defects in concrete. In this area, \citet{Yang2022-nl} utilized a gradient boosting decision tree (GBDT) ensemble model to effectively classify defect signals from embedded piezoelectric sensors. Further supporting these findings, a review on ultrasonic weld inspection by \citet{Sun2023-yp} affirms the wide applicability of machine learning, while also stressing that successful outcomes depend on careful data selection and precise hyperparameter tuning.

In addition to specific uses, researchers have explored the significant impact of ML on the wider fields of NDI and structural health monitoring (SHM). A study by \citet{Gardner2020-bn} demonstrated that ML improves ultrasonic NDI across several functions, including data processing, damage detection, localization, and prognosis. The same study also identified persistent challenges related to data quality, quantity, and complexity that can hinder reliable results. Building on this premise, \citet{Taheri2022-ix} reviewed the use of different learning approaches such as supervised, unsupervised, and reinforcement learning for common tasks like feature extraction and classification. They further suggested that integrating ML with smart technologies, particularly digital twins that create real time simulations of physical assets, could optimize the entire NDI process through predictive maintenance and early anomaly detection.

Demonstrating the broad applicability of these techniques, ML has also found significant use in diverse industries such as food quality assessment. A comprehensive review by \citet{Lin2023-fd} explains how both traditional and deep learning models can be paired with novel non-destructive methods like spectral imaging and acoustic analysis. This combination has led to successful applications for variety classification, contamination detection, and spoilage identification. However, mirroring the challenges found in structural analysis, the authors also note that issues with data availability and interoperability remain significant hurdles to widespread adoption.

\subsection{Deep Learning} \label{sec:review-deep-learning}

We distinguish DL from traditional ML in this review to highlight its superior capacity for modeling the high-dimensional, unstructured data typical of modern NDI. Fueled by the massive increase in computational resources, DL has solidified its position as a transformative tool within the research community, consistently yielding high accuracy in complex inspection tasks. As a result, numerous studies have been dedicated to proposing novel deep learning models \citep{Akcay2019-xy, Ma2020-cu, Siegel2020-cp, Presenti2022-kp} and systematically evaluating their performance against established benchmarks \citep{Yousefi2018-el, Posilovic2022-ay} to validate their reliability.

Much of the initial research focused on validating the effectiveness of various DL architectures for NDI. For instance, an early study by \citet{Yousefi2018-el} successfully applied a deep learning approach to identify defects in infrared thermography (IRT) images from both steel and carbon fiber reinforced plastic specimens (CFRP), demonstrating promising performance. More recent comparative analysis by \citet{Posilovic2022-ay} reviewed several DL models for anomaly detection in ultrasonic scans, finding that a modified PaDiM model outperformed others like Ganomaly and DifferNet on their dataset by successfully identifying all anomalies. Similarly, a broad review by \citet{Sun2023-yp} on ultrasonic weld inspection affirmed the general utility of machine learning but stressed that success is contingent upon meticulous data selection and hyperparameter tuning.

Beyond standard benchmarking, significant innovation has been directed toward addressing specific NDI challenges such as data scarcity. Semi-supervised models have emerged as a powerful solution to this problem. \citet{Akcay2019-xy} introduced the GANomaly model, which utilizes a conditional generative adversarial network to achieve high detection accuracy using minimal labeled data. Adopting a similar reconstruction-based strategy, \citet{Presenti2022-kp} developed an autoencoder method for X-ray analysis that learns to reconstruct non-defective objects. By using the reconstruction error as a distinct feature, this model classifies defective parts with high accuracy while maintaining robustness against noise.

Deep learning has also been tailored to a wide array of specialized applications. \citet{Ma2020-cu} developed a CNN model to predict seed viability from near infrared spectral images, achieving over $90\%$ classification accuracy. In advanced manufacturing, a deep neural network proposed by \citet{Siegel2020-cp} proved capable of detecting delamination defects as small as $0.1mm$ in $3D$ printed components, far surpassing conventional methods. Demonstrating efficiency with small datasets, \citet{Na2021-eq} combined an electromechanical impedance technique with a probabilistic neural network to identify bolt loosening with over $94\%$ accuracy from only $20$ data points. Recent advancements continue to push these boundaries, with \citet{Tabernik2023-dp} developing segmentation models for pavement surface defects that optimize both performance and computational speed. Furthermore, \citet{Barshok2025-be} introduced CATT-S, a model that captures inter-beam dependencies in phased array ultrasonic testing to detect defects in complex structures with $99.4\%$ accuracy.

\subsection{Anomaly Detection} \label{sec:review-anomaly-detection}

Anomaly detection represents a critical methodology in NDI, defined by its objective to identify data points or regions that deviate significantly from an established norm. While this domain frequently employs DL architectures, it is distinct from traditional supervised learning approaches. Whereas standard DL models are typically trained to classify specific, known defect types based on labeled examples, anomaly detection strategies focus on learning a robust representation of normal or healthy conditions. This shift in paradigm, from defect recognition to deviation detection, is particularly valuable in the NDI context, where obtaining large, annotated datasets of diverse defects is often impractical compared to the abundance of data available from non-defective components.

A prominent strategy in this field involves training DL models on exclusively normal or non-defective data. For example, the GANomaly method introduced by \citet{Akcay2019-xy} uses a generative adversarial network in a semi-supervised fashion. The model's generator learns the underlying distribution of normal training data, which enables the discriminator to effectively identify any inputs that do not conform to this learned standard. This technique achieves high accuracy with minimal labeled data, outperforming many traditional methods. A similar principle was applied by \citet{Presenti2022-kp}, who presented an autoencoder-based method for finding anomalies in X-ray images. Their model learns to reconstruct images of non-defective objects, and the magnitude of the reconstruction error is then used to classify an object as either normal or defective, proving to be both highly accurate and tolerant to noise.

Other approaches apply deep learning models more directly to the task of defect identification. A study by \citet{Baumgartl2020-fe} proposed using a convolutional neural network to find defects during the additive manufacturing process using thermographic imaging data. Their approach successfully achieved high detection accuracy while maintaining low computational costs compared to other state-of-the-art models.

The trends and challenges in NDI anomaly detection reflect a broader conversation in the field of computer science. A comprehensive review by \citet{Alloqmani2021-gz} of anomaly detection methods in general and medical imaging found that deep learning models are the most popular and frequently used. The authors highlight a key advantage of these models, which is their ability to learn complex patterns from large datasets without manual feature engineering. However, they also identify significant challenges, including the scarcity of labeled anomaly data, the difficulty of establishing clear benchmarks, and the complex task of selecting an appropriate model architecture. This analysis underscores the great potential of deep learning for anomaly detection in NDI while acknowledging the challenges that lie ahead.

\subsection{Generative Model} \label{sec:review-generative-network}

Generative models, particularly Generative Adversarial Networks (GANs), have emerged as a powerful and flexible tool in NDI for tasks ranging from anomaly detection to data augmentation. These models excel at learning the underlying distribution of complex data, which allows them to either generate new, realistic data samples or identify deviations from a learned norm. These models warrant a dedicated discussion within this review due to their unique capacity to synthesize realistic inspection data, offering a powerful computational solution to the chronic shortage of labeled defect samples that typically hinders the training of robust DL algorithms.

A primary application of this technology is semi-supervised anomaly detection, where models are trained on normal data to find defects. A notable example is the GANomaly method proposed by \citet{Akcay2019-xy}. This approach uses a conditional GAN where a generator learns to create realistic images of non-defective components. A discriminator then distinguishes between real and generated images, effectively learning the standard for normal data. The evaluation of GANomaly on various datasets proved it highly effective, outperforming traditional methods by achieving high accuracy with very little labeled data. Following a similar concept, \citet{Hashimoto2021-mf} used a deep generative model to identify small cracks in magneto-optical images of steel specimens, successfully detecting slit defects just 0.15 mm wide.

Beyond anomaly detection, generative models are increasingly used for data augmentation to improve the performance of other machine learning models. A study by \citet{Fang2021-nw} demonstrates this by combining a dimension reduction technique with a GAN to augment thermal images. Their method successfully enhanced the visibility of defects and expanded the diversity of the original image dataset, leading to promising results.

Generative models have also been applied to more specialized predictive tasks. For instance, \citet{Angermann2023-at} developed a method to inspect cylinder liners in large engines by using a GAN to predict surface texture from microscope images. Their experiment accurately predicted surface texture depth in the micrometer range, proving the method's viability for detailed inspections. However, the authors noted that its sensitivity to external factors like vibration currently limits its use for on-site applications.

\section{Techniques of NDI} \label{sec:review-techniques-of-ndi}

\begin{figure}[pos=htp]
    \centering
    \includegraphics[width=1\linewidth]{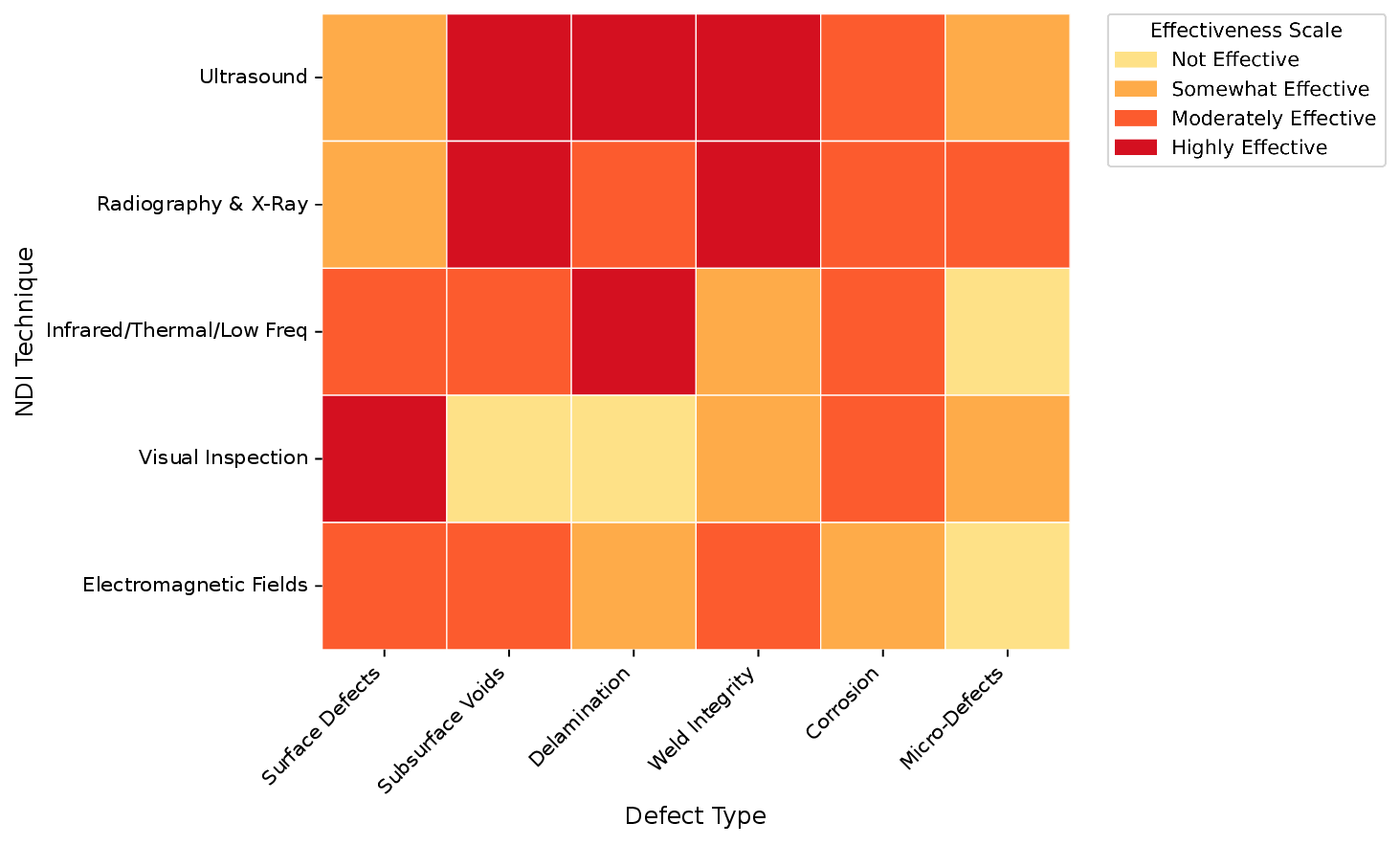}
    \caption{Effectiveness of different NDI techniques.}
    \label{fig:review-ndi-effectiveness-of-techniques}
\end{figure}

The techniques used for NDI vary based on the individual industry and the task at hand. Easy deployment and cost-effectiveness also play an essential role in this case. There are a variety of techniques in modern NDI, including but not limited to ultrasound, radiography, thermal, laser, and many more. This section discusses the studies regarding various techniques used for NDI. To illustrate the distribution of research across different physical modalities, Table \ref{tab:review-techniques-paper-domain} classifies the reviewed papers according to the specific inspection technologies employed, highlighting the prevalence of ultrasonic and infrared-thermal methods. An overview of the effectiveness of different NDI techniques is visualized in Figure \ref{fig:review-ndi-effectiveness-of-techniques}.

\begin{longtable}[c]{lp{0.8cm}p{0.8cm}p{0.8cm}p{0.8cm}p{0.8cm}}
    \caption{Domains of the research papers discussed in Section \ref{sec:review-techniques-of-ndi}. Domain labels are U = Ultrasound, RX = Radiography and X-Ray, ITLF = Infrared, Thermal, and Low Frequency, VI = Visual Inspection, and EMF = Electromagnetic Field.\label{tab:review-techniques-paper-domain}}\\
    \toprule
         Paper & U & RX & ITLF & VI & EMF \\
    \midrule
    \endfirsthead
    \toprule
         Paper & U & RX & ITLF & VI & EMF \\
    \midrule
    \endhead
         \citet{Cawley1984-le} & & & \cellcolor{cellbg} & & \\
         \citet{Cawley1994-uf} & & & \cellcolor{cellbg} & & \\
         \citet{Lange1994-nf} & & & & & \cellcolor{cellbg} \\
         \citet{Ould_Naffa2002-ab} & \cellcolor{cellbg} & & & & \\
         \citet{Han2001-ch} & & & \cellcolor{cellbg} & & \\
         \citet{Huang2003-gy} & & \cellcolor{cellbg} & & & \\
         \citet{Ruzek2006-pq} & \cellcolor{cellbg} & & & \cellcolor{cellbg} & \\
         \citet{Kosmas2007-uf} & & & & & \cellcolor{cellbg} \\
         \citet{Wang2008-hg} & \cellcolor{cellbg} & & & & \\
         \citet{Yang2013-mh} & \cellcolor{cellbg} & & & & \\
         \citet{Na2016-vf} & & & \cellcolor{cellbg} & & \\
         \citet{Dwivedi2018-ts} & \cellcolor{cellbg} & & & & \\
         \citet{Yousefi2018-el} & & & \cellcolor{cellbg} & & \\
         \citet{Zaporozhets2019-cn} & & & \cellcolor{cellbg} & & \\
         \citet{Al_Rahman_M_Abu_Ebayyeh2020-ox} & & & & \cellcolor{cellbg} & \\
         \citet{El_Masri2020-ux} & \cellcolor{cellbg} & & & & \\
         \citet{Gardner2020-bn} & \cellcolor{cellbg} & & & & \\
         \citet{Honarvar2020-tw} & \cellcolor{cellbg} & & & & \\
         \citet{Ma2020-cu} & & & \cellcolor{cellbg} & & \\
         \citet{Mishra2020-qv} & \cellcolor{cellbg} & & & & \\
         \citet{Siegel2020-cp} & & & \cellcolor{cellbg} & & \\
         \citet{Sreeshan2020-dc} & & & \cellcolor{cellbg} & & \\
         \citet{Wang2020-up} & \cellcolor{cellbg} & \cellcolor{cellbg} & \cellcolor{cellbg} & & \\
         \citet{Zhang2020-kp} & \cellcolor{cellbg} & \cellcolor{cellbg} & & & \\
         \citet{Deepak2021-pa} & \cellcolor{cellbg} & \cellcolor{cellbg} & & \cellcolor{cellbg} & \cellcolor{cellbg} \\
         \citet{Fernandes2021-sf} & \cellcolor{cellbg} & \cellcolor{cellbg} & \cellcolor{cellbg} & & \\
         \citet{Khedmatgozar_Dolati2021-as} & \cellcolor{cellbg} & \cellcolor{cellbg} & \cellcolor{cellbg} & \cellcolor{cellbg} & \cellcolor{cellbg} \\
         \citet{Na2021-eq} & & & & & \cellcolor{cellbg} \\
         \citet{Nsengiyumva2021-bo} & \cellcolor{cellbg} & \cellcolor{cellbg} & \cellcolor{cellbg} & \cellcolor{cellbg} & \cellcolor{cellbg} \\
         \citet{Song2021-ir} & & & & & \cellcolor{cellbg} \\
         \citet{Gupta2022-id} & \cellcolor{cellbg} & \cellcolor{cellbg} & & \cellcolor{cellbg} & \cellcolor{cellbg} \\
         \citet{Mahanti2022-sp} & & \cellcolor{cellbg} & \cellcolor{cellbg} & & \\
         \citet{Posilovic2022-ay} & \cellcolor{cellbg} & & & & \\
         \citet{Presenti2022-kp} & & \cellcolor{cellbg} & & & \\
         \citet{Yang2022-nl} & & & \cellcolor{cellbg} & & \\
         \citet{Bajgholi2023-sr} & \cellcolor{cellbg} & \cellcolor{cellbg} & & & \\
         \citet{Jodhani2023-mn} & \cellcolor{cellbg} & & & & \\
         \citet{Sun2023-yp} & \cellcolor{cellbg} & & & & \\
         \citet{Barshok2025-be} & \cellcolor{cellbg} & & & & \\
         \citet{Rahman2025-wf} & \cellcolor{cellbg} & & & & \\
         \citet{Ribeiro2025-mf} & & & & \cellcolor{cellbg} & \\
    \bottomrule
    
\end{longtable}

\subsection{Ultrasound} \label{sec:review-ultrasound}

UT stands as one of the most widely utilized techniques in the field of NDI. Its significance is reflected in the extensive academic literature dedicated to its evolution and enhancement. Comprehensive reviews by \citet{Dwivedi2018-ts}, \citet{El_Masri2020-ux}, \citet{Wang2020-up}, \citet{Nsengiyumva2021-bo}, and \citet{Gupta2022-id} establish UT as a valuable method for identifying subsurface defects in materials that conduct sound. However, these studies also highlight the limitations, which include difficulty detecting very small defects \citep{Nsengiyumva2021-bo}, a reliance on the skill of the inspector, and the potential for misclassifying flaws. The choice of a specific ultrasonic technique is therefore highly dependent on the requirements of the application, as noted by \citet{Deepak2021-pa} in the context of steel weld joints.

The versatility of UT is evident in its application across a diverse range of industries. Review by \citet{El_Masri2020-ux} highlighted the diverse advantages, including equipment mobility, surface penetration ability, physical defect detection, and so on. \citet{Honarvar2020-tw} reviewed its use in additive manufacturing, underscoring its value and outlining future research directions to address challenges related to geometry and defect size. Similarly, its utility has been recognized in structural health monitoring \citep{Khedmatgozar_Dolati2021-as} and for inspecting composite materials \citep{Ruzek2006-pq, Jodhani2023-mn}. In a direct comparison, \citet{Ruzek2006-pq} found that for certain impact damage, other methods like shearography could outperform ultrasonic C-scan, further emphasizing the need for careful technique selection.

A significant body of research is dedicated to evaluating and improving ultrasonic methods. Studies have tested UT on various materials to gauge its efficiency, with \citet{Ould_Naffa2002-ab} finding that surface waves were the most promising among the three types of waves, longitudinal, transverse, and surface waves. Similarly, \citet{Bajgholi2023-sr} evaluated the performance of UT for inspecting weld joints in hydraulic turbine runners and concluded that phased array ultrasonic testing (PAUT) using the total focusing method (TFM) outperforms conventional UT. Researchers have also proposed novel ways to enhance UT. For instance, \citet{Yang2013-mh} successfully used ultrasonic thermography to identify the shape and depth of impact damage on aerospace components. To overcome challenges like fiber waviness in composites, which can obscure defects, \citet{Zhang2020-kp} proposed a combined analysis of amplitude and B-scan images. Technological advancements have also led to new equipment, such as the laser ultrasonic visualization system presented by \citet{Wang2008-hg}, which allows for imaging wave propagation in complex objects. Furthermore, \citet{Fernandes2021-sf} demonstrated that combining air-coupled ultrasound with other NDT techniques like X-ray micro tomography and infrared thermography provides a more comprehensive characterization of complex objects like ancient marquetry.

Looking toward the future, the integration of artificial intelligence and automation is seen as pivotal for advancing UT, a point underscored by \citet{Jodhani2023-mn}. The use of ML and DL in ultrasonic NDI has been reviewed by several authors \citep{Gardner2020-bn, Mishra2020-qv, Posilovic2022-ay, Sun2023-yp}. \citet{Gardner2020-bn} discussed how machine learning can improve data processing, damage detection, and localization, while also noting challenges related to data quality and complexity. A review by \citet{Sun2023-yp} on weld inspection affirmed the broad utility of ML but stressed that success requires careful data selection and hyperparameter tuning, underscoring the need for common datasets to advance the field. Specific applications have shown great promise as well. \citet{Mishra2020-qv} evaluated several models for predicting the compressive strength of historic masonry, finding that a support vector regression model offered a practical and promising solution for UT. For anomaly detection, \citet{Posilovic2022-ay} found that a modified PaDiM deep learning model performed best in their tests on ultrasonic scans, identifying all anomalies successfully. Recent innovations have further refined these capabilities. \citet{Barshok2025-be} introduced the CATT-S model, achieving $99.4\%$ accuracy by modeling inter-beam dependencies in phased array data, while \citet{Rahman2025-wf} demonstrated the efficacy of instance segmentation architectures like Mask R-CNN and YOLO for parsing complex scans of aerospace composites.

\subsection{Radiography and X-Ray} \label{sec:review-radiography-and-x-ray}

Radiography and X-ray techniques are highly valued in NDI for their exceptional ability to detect subsurface defects. The capabilities, limitations, and ongoing potential of these technologies have been the subject of extensive academic discussion and research.

Several comprehensive reviews have established the role of radiography within the broader landscape of NDI methods. Studies by \citet{Wang2020-up}, \citet{Nsengiyumva2021-bo}, \citet{Gupta2022-id}, and \citet{Bajgholi2023-sr} highlight the effectiveness of radiography for subsurface flaw detection in materials like weld joints and composites. However, they also point out significant limitations, including safety hazards, the challenge of waste disposal, high costs, relatively slow inspection, and strict orientation requirements for successful inspection. The consensus, as noted by \citet{Deepak2021-pa} and \citet{Khedmatgozar_Dolati2021-as}, is that while radiography has its strengths, the optimal NDI technique ultimately depends on the specific demands of the application, whether it be for steel weld joints or local damage detection on bridges.

The application of X-ray technology, particularly computed tomography (CT), is widespread across various industries. In industrial settings, visualizing high resolution CT data can be challenging due to noise. \citet{Huang2003-gy} addressed this by developing a method combining region growing techniques with a 2D histogram interface, enabling clearer feature extraction and comparison within volumetric data. The versatility of X-ray imaging is further demonstrated in its use for cultural heritage, where \citet{Fernandes2021-sf} used X-ray micro tomography to create detailed three-dimensional images of an ancient marquetry sample as part of a multi-technique characterization. In the aerospace industry, \citet{Zhang2020-kp} utilized X-ray CT alongside ultrasonic scans to analyze composite samples, discovering how fiber waviness could impede other inspection methods. X-ray imaging is also valuable in the food industry, where it is used for damage detection, including finding bruises, cracks, and internal defects in fruit, ensuring consumer safety \citep{Mahanti2022-sp}.

Modern advancements have integrated ML with radiographic analysis to improve accuracy and automation. A notable example is the anomaly detection method proposed by \citet{Presenti2022-kp}. Their approach uses an autoencoder trained on images of non-defective objects. When presented with a new object, the model's ability to reconstruct the image is used to determine if it is defective. This method proved highly accurate and noise-tolerant on a simulated radiograph dataset, showcasing the potential of DL to enhance radiographic inspection.

\subsection{Infrared, Thermal, and Low Frequency} \label{sec:review-low-frequency}

Techniques involving infrared, thermal, and low frequency signals represent a versatile and powerful group of NDI methods. These approaches, including impedance testing, transient thermography, and shearography, have been widely adopted for their speed and effectiveness in detecting a range of subsurface defects like delaminations and disbonds.

Foundational research established the utility of these methods. For instance, the impedance method was recognized early for its speed and the convenience of allowing dry surface contact during testing \citep{Cawley1984-le}. Over time, technological advancements, especially in thermal imaging, made these NDI applications faster and more accessible \citep{Cawley1994-uf}. Innovative combinations of techniques also emerged, such as "thermosonics," which pairs ultrasonic excitation with infrared imaging to effectively detect adhesion defects and other damages \citep{Han2001-ch}.

The application of these techniques spans a wide variety of industries and environments. In structural health monitoring, particularly for infrastructure like steel bridges, these advanced NDI methods are critical for local damage detection \citep{Khedmatgozar_Dolati2021-as}. Research has also explored novel deployment strategies, such as attaching piezoelectric sensors to unmanned aerial vehicles to inspect structures without the need for expensive image processing \citep{Na2016-vf}. The analysis methods have also been refined, with techniques like the Hilbert transform being used to extract more effective features from impedance signals in composite structures \citep{Zaporozhets2019-cn}.

In advanced manufacturing and aerospace, active thermography is used to find delamination defects in 3D printed components \citep{Siegel2020-cp}, while pulsed thermography is a proven method for detecting flaws in aerospace composite laminates \citep{Sreeshan2020-dc}. These techniques are frequently reviewed for their effectiveness in inspecting composite materials \citep{Wang2020-up}. Their utility extends to other fields as well, including cultural heritage, where infrared thermography helps characterize ancient marquetry samples \citep{Fernandes2021-sf}, and the food industry, where thermal and near infrared imaging are used for tasks like fruit damage assessment and seed viability prediction \citep{Mahanti2022-sp, Ma2020-cu}. However, a key drawback of these methods is the inability to detect the precise shape and depth of the defect \citep{Wang2020-up, Nsengiyumva2021-bo}.

The integration of ML and DL has significantly enhanced the capabilities of these NDI methods. Researchers have successfully applied DL for defect detection in infrared thermography images by using pretrained convolutional neural networks to identify subtle temperature variations \citep{Yousefi2018-el}. In the case of 3D printed parts, a deep neural network was trained on thermographic data to detect, as well as determine the severity of delaminations \citep{Siegel2020-cp}. For agricultural applications, a combination of ML classifiers and DL models has been used to achieve high accuracy in predicting seed viability from near infrared data \citep{Ma2020-cu}. Furthermore, ensemble ML models are being used to classify signals from embedded piezoelectric sensors to evaluate subsurface defects in concrete \citep{Yang2022-nl}.

\subsection{Visual Inspection} \label{sec:review-visual-inspection}

Visual inspection is a foundational NDI technique, though its modern applications are often automated and compared against other methods. This approach is desirable due to its fast, low-cost, and non-contact characteristics \citep{Nsengiyumva2021-bo}. Several reviews have evaluated its role in various industrial contexts. For instance, \citet{Deepak2021-pa} reviewed its use for inspecting low-carbon steel weld joints alongside four other NDI techniques. The authors concluded that the best method depends entirely on the application's specific requirements, noting that laser beam welding proved highly acceptable across multiple NDI methods. This perspective is shared by \citet{Gupta2022-id}, who, after reviewing numerous NDI techniques, concluded that while the field is mature, no single method is suitable for all situations and that combining techniques often produces better results. Similarly, a review focused on local damage detection on steel bridges emphasized the need for advanced NDI techniques, outlining the distinct advantages and disadvantages of existing methods like visual inspection \citep{Khedmatgozar_Dolati2021-as}. One major disadvantage is the inability to detect sub-surface defects \citep{Nsengiyumva2021-bo}.

While common, the effectiveness of manual visual inspection can be limited compared to more advanced technologies. In a direct comparison of methods for identifying impact damage, \citet{Ruzek2006-pq} tested visual inspection, ultrasonic C-scan, and shearography, finding that shearography performed best in that specific case. To address such limitations and improve reliability, the field has increasingly moved toward automated optical inspection (AOI). A review of AOI techniques within the electronics manufacturing industry discussed the associated processes, technologies, and defect identification capabilities \citep{Al_Rahman_M_Abu_Ebayyeh2020-ox}. The authors highlighted that the quality of the hardware and software setup is vital for the acquired image quality and, consequently, the inspection results. They also identified future research directions, such as overcoming the significant limitation of detecting subsurface defects with optical methods. The practical viability of these automated systems continues to expand into new sectors, as recently demonstrated by \citet{Ribeiro2025-mf}, who successfully implemented automated visual defect detection for pharmaceutical bottle production.

\subsection{Electromagnetic Field} \label{sec:review-electromagnetic-field}

Inspection methods based on electromagnetic fields, along with impedance analysis, form another important category of NDI techniques. General reviews of NDI methods have frequently discussed the role of techniques like magnetic particle testing for specific applications such as inspecting low carbon steel weld joints \citep{Deepak2021-pa} or performing local damage detection on steel bridges \citep{Khedmatgozar_Dolati2021-as}. These reviews often conclude that while no single technique is universally suitable, magnetic testing has its distinct advantages \citep{Gupta2022-id}. One such advantage, highlighted in a review of mechanical impedance analysis, is the ability to use dry point contact, making the method applicable to almost any material \citep{Lange1994-nf}.

Research in this area has focused on developing novel techniques and improving existing ones. For example, \citet{Kosmas2007-uf} proposed a hybrid method that combines magnetic anomaly detection with eddy current testing, which successfully detected subsurface defects in ferromagnetic materials. Efforts have also been made to solve persistent challenges like signal noise, with \citet{Song2021-ir} proposing an electromagnetic induction impedance method specifically to improve the inspection of grounding grids at power stations. Furthermore, these techniques have been successfully integrated with machine learning. \citet{Na2021-eq} combined an electromechanical impedance-based technique with a probabilistic neural network to accurately identify bolt loosening, achieving high performance even with a very small dataset. The major drawback for wider adoption of this method is the dependence on conductive materials \citep{Nsengiyumva2021-bo}.

\section{Application Areas} \label{sec:review-application-areas}
NDI plays a pivotal role in ensuring the safety, reliability, and longevity of critical infrastructure and manufactured components across a diverse spectrum of industries. While the fundamental principles of defect detection remain consistent, the practical application of NDI is heavily dictated by the unique material properties, geometric complexities, and safety standards of each sector. From the high-stakes environment of aerospace composites to the delicate preservation of cultural heritage artifacts, each domain presents a distinct set of challenges that drive specialized research and technological adaptation. This section examines the deployment of NDI across key industries, highlighting how sector-specific requirements have shaped the evolution and integration of inspection technologies. Table \ref{tab:review-application-paper-domain} categorizes the literature by industrial sector, offering a clear view of how NDI advancements are being tailored to meet the distinct material and safety requirements of fields such as aerospace, civil infrastructure, and heritage conservation.

\begin{longtable}[c]{lp{0.8cm}p{0.8cm}p{0.8cm}p{0.8cm}p{0.8cm}p{0.8cm}}
    \caption{Domains of the research papers discussed in Section \ref{sec:review-application-areas}. Domain labels are CM = Composite materials, OMFS = Organic Materials and Food Safety, CnM = Construction Materials, FA = Fragile Artifacts, M = Medical, and SHM = Structural Health Monitoring.\label{tab:review-application-paper-domain}}\\
    \toprule
         Paper & CM & OMFS & CnM & FA & M & SHM \\
    \midrule
    \endfirsthead
    \toprule
         Paper & CM & OMFS & CnM & FA & M & SHM \\
    \midrule
    \endhead
         \citet{Nemet1997-va} & & & & & \cellcolor{cellbg} & \\
         \citet{Ould_Naffa2002-ab} & & & \cellcolor{cellbg} & & & \\
         \citet{Brosnan2004-yd} & & \cellcolor{cellbg} & & & & \\
         \citet{Diamanti2005-mr} & \cellcolor{cellbg} & & & & & \\
         \citet{Farrar2007-jo} & & & & & & \cellcolor{cellbg} \\
         \citet{Achenbach2009-jj} & & & & & & \cellcolor{cellbg} \\
         \citet{Yang2013-mh} & \cellcolor{cellbg} & & & & & \\
         \citet{Lee2014-xt} & & & & & & \cellcolor{cellbg} \\
         \citet{Martinho2014-kh} & & & & \cellcolor{cellbg} & & \\
         \citet{Memmolo2015-er} & & & & \cellcolor{cellbg} & & \\
         \citet{Du_Plessis2016-qn} & & & & & \cellcolor{cellbg} & \\
         \citet{Gholizadeh2016-ww} & \cellcolor{cellbg} & & & & & \\
         \citet{Na2016-vf} & & & & & & \cellcolor{cellbg} \\
         \citet{Dwivedi2018-ts} & \cellcolor{cellbg} & & & & & \\
         \citet{Sreeshan2019-sf} & \cellcolor{cellbg} & & & & & \\
         \citet{Zaporozhets2019-cn} & \cellcolor{cellbg} & & & & & \\
         \citet{Abu_Dabous2020-qp} & & & & & & \cellcolor{cellbg} \\
         \citet{Gardner2020-bn} & & & & \cellcolor{cellbg} & & \\
         \citet{Guemes2020-yl} & & & & & & \cellcolor{cellbg} \\
         \citet{Honarvar2020-tw} & \cellcolor{cellbg} & & & & & \\
         \citet{Ma2020-cu} & & \cellcolor{cellbg} & & & & \\
         \citet{Mishra2020-qv} & & & & \cellcolor{cellbg} & & \\
         \citet{Siegel2020-cp} & \cellcolor{cellbg} & & & & & \\
         \citet{Sreeshan2020-dc} & \cellcolor{cellbg} & & & & & \\
         \citet{Deepak2021-pa} & & & \cellcolor{cellbg} & & & \\
         \citet{Khedmatgozar_Dolati2021-as} & & & & & & \cellcolor{cellbg} \\
         \citet{Kot2021-xe} & & & & & & \cellcolor{cellbg} \\
         \citet{Mahanti2022-sp} & & \cellcolor{cellbg} & & & & \\
         \citet{Yang2022-nl} & & & & & & \cellcolor{cellbg} \\
         \citet{Hola2023-sh} & & & & & & \cellcolor{cellbg} \\
         \citet{Jodhani2023-mn} & \cellcolor{cellbg} & & & & & \\
         \citet{Hassani2023-hi} & & & & & & \cellcolor{cellbg} \\
         \citet{Lin2023-fd} & & \cellcolor{cellbg} & & & & \\
    \bottomrule
    
\end{longtable}

\subsection{Composite Materials} \label{sec:review-composite-materials}

Composite materials are extensively used in aerospace, manufacturing, and civil infrastructure, making their inspection a critical application area for NDI \citep{Dwivedi2018-ts}. Due to their complex structures, a wide variety of NDI methods are employed to evaluate their integrity. A comprehensive review by \citet{Gholizadeh2016-ww} provided an overview of seven common NDI methods for composites, including visual testing, ultrasonics, thermography, and shearography, highlighting that the correct method must be chosen based on the material, the defect type, and the cost-benefit ratio.

Ultrasonic testing is a frequently applied technique. A review by \citet{Jodhani2023-mn} grouped ultrasonic methods for composites into conventional and non-conventional categories, concluding that the future of this field will heavily depend on automation and artificial intelligence. \citet{Honarvar2020-tw} also reviewed ultrasonic techniques in the related field of additive manufacturing, listing current challenges. Specific ultrasonic approaches have also been developed, such as using low-frequency Lamb waves to identify hairline defects as small as $3mm$ in repaired composite patches, offering a cost-saving method for in-field testing \citep{Diamanti2005-mr}.

Thermal and thermographic techniques are also highly effective. \citet{Yang2013-mh} used ultrasonic thermography to successfully identify the shape and depth of impact damage on aerospace components. For 3D printed components, \citet{Siegel2020-cp} presented a novel method using active thermography and a deep neural network to detect delamination defects as small as $0.1mm$, an approach that significantly outperforms conventional methods. Research has also focused on enhancing thermographic image processing. \citet{Sreeshan2019-sf} used Gabor filters and watershed segmentation to improve the contrast of debonding defects. In a subsequent study, \citet{Sreeshan2020-dc} developed a full pipeline for pulsed thermography, including pre-processing and segmentation, which successfully identified the majority of simulated defects in an aerospace composite laminate.

Other NDI methods have been adapted for composite structures as well. For instance, \citet{Zaporozhets2019-cn} proposed using the Hilbert transform to improve the analysis of impedance method signals. They found that features extracted using this mathematical technique were more effective for defect detection.

\subsection{Organic Materials and Food Safety} \label{sec:review-organic-materials}

The inspection of organic materials, particularly in the food industry, relies heavily on non-destructive techniques to ensure quality and consumer safety. Computer vision is a key method in this area, offering speed, consistency, and objectivity for tasks like defect detection, grading, and sorting \citep{Brosnan2004-yd}. As the authors note, the success of these automated systems is highly dependent on a proper illumination and image acquisition setup.

Beyond standard computer vision, other NDI technologies are also employed. \citet{Mahanti2022-sp} discussed the use of biospeckle, X-ray imaging, hyperspectral imaging (HSI), and thermal imaging (TI) for detecting fruit damage. These non-destructive methods are advantageous because they allow for online inspection and real-time damage assessment without harming the product. The authors presented successful applications for finding various damages, including bruises, cracks, internal defects, and insect infestation.

The effectiveness of these NDI technologies is greatly amplified by machine learning. \citet{Lin2023-fd} provided a comprehensive overview of both traditional and deep learning techniques for food quality testing. They underscored the promising results from combining novel technologies like acoustic analysis, machine vision, and spectral imaging with machine learning algorithms. These combinations have been successfully applied to tasks such as variety classification, contamination detection, and spoilage detection, although challenges related to data availability and quality remain.

A specific example of this integration is the deep learning method for seed viability prediction proposed by \citet{Ma2020-cu}. Their approach used near-infrared hyperspectral imaging to collect spectral data from seeds. After preprocessing the data and using machine learning for feature extraction, a convolutional neural network (CNN) was constructed to automatically identify the differences between viable and non-viable seeds. The CNN model achieved a high classification accuracy of $90.4\%$ on a fivefold cross-validation set, proving it to be a promising tool for rapid and non-destructive seed viability prediction.

\subsection{Construction Materials} \label{sec:review-construction-materials}

NDI is essential for ensuring the safety and integrity of various construction materials, from concrete structures to steel components. Research in this area has focused on both developing specific applications and evaluating the effectiveness of different NDI techniques for common materials.

In the inspection of concrete, for example, \citet{Ould_Naffa2002-ab} explored the use of high-frequency ultrasound to identify chemical damage. Their study compared three types of waves and found that surface waves were the most promising for this application.

For metallic components, such as low-carbon steel weld joints, a comparative review by \citet{Deepak2021-pa} evaluated five different NDT techniques, including visual inspection, radiography, and ultrasonic testing. The authors tested these methods on joints created by three types of welding. They concluded that the best NDI technique depends on the specific requirements of the application, noting that the laser beam-welded plate was highly acceptable across multiple NDI methods.

\subsection{Fragile Artifacts} \label{sec:review-fragile-artifacts}

NDI is particularly valuable for the conservation and restoration of fragile artifacts and cultural heritage. Geophysical methods are one set of tools used for this purpose. A review by \citet{Martinho2014-kh} discussed the use of techniques such as ground penetrating radar, seismic methods, and electrical resistivity tomography to gather information about subsurface structures in heritage sites. These methods are useful for identifying voids, cracks, moisture content, and other characteristics, making them valuable tools for conservation and restoration of heritage structures.

Advanced signal processing is also adapted for these sensitive applications. For instance, \citet{Memmolo2015-er} proposed an automated process to analyze Digital Speckle Pattern Interferometry signals for art restoration. Their method filters external noise by analyzing Fourier spectrum frequencies and image sharpness, which helps experts monitor artifacts like wooden paintings in noisy environments.

Furthermore, ML is increasingly being applied to data gathered from historic structures. \citet{Mishra2020-qv} evaluated three ML models to predict the compressive strength of unreinforced brick masonry in buildings such as museums. Using ultrasonic pulse velocity and rebound hammer numbers as inputs, they found that a support vector regression model outperformed both artificial neural networks and adaptive neuro fuzzy inference systems, offering a practical solution for strength prediction. This aligns with a broader discussion by \citet{Gardner2020-bn} on the positive impact of ML in ultrasonic non-destructive evaluation and structural health monitoring. They claim ML can improve data processing, damage detection, localization, and prognosis. Despite challenges like data quality and quantity, the authors affirm that ML is a powerful tool for enhancing NDI and SHM systems.

\subsection{Medical} \label{sec:review-medical}
The practices of non-invasive testing in the medical field and non-destructive inspection in manufacturing share significant parallels, with both domains frequently learning from one another's advancements \citep{Nemet1997-va}. Observing the technologies used in each field is therefore imperative for inspiring future NDI techniques. A clear example of this crossover is found in the work of \citet{Du_Plessis2016-qn}, who compared the performance of medical and industrial CT scanners. The authors explored the potential of using medical CT scanners for industrial non-destructive inspection. They acknowledged the limitations of medical systems, such as comparatively lower resolution and penetration depth. However, they also noted that medical CT scanners can scan large objects and offer higher throughput. This led them to propose that medical CT systems can be a viable alternative for industrial applications where extremely high resolution is not critical, potentially reducing both inspection costs and time significantly.

\subsection{Structural Health Monitoring} \label{sec:review-structural-health-monitoring}
SHM is a critical application area for NDI, focused on ensuring the safety, reliability, and longevity of infrastructure. A comprehensive SHM system involves several steps, but its implementation faces significant challenges, including structural degradation, a lack of quality data for defect classification, and convincing the industry of the cost benefit \citep{Farrar2007-jo}. Despite these hurdles, the potential benefits of SHM, such as increased safety and reduced maintenance costs, are substantial \citep{Achenbach2009-jj}.

The application of NDI techniques is fundamental to SHM, particularly for inspecting large structures like bridges. \citet{Lee2014-xt} reviewed bridge inspection practices in the U.S., classifying NDT methods for concrete and steel components and noting that few states explicitly leverage their full potential. The need for advanced NDI techniques in SHM, especially for local damage detection on steel bridges, has been strongly emphasized \citep{Khedmatgozar_Dolati2021-as}.

A wide array of NDI techniques is employed in SHM. Reviews in the field have covered methods including ground penetrating radar, infrared thermography, fiber optic sensors, camera-based methods, laser scanners, acoustic emissions, and ultrasound \citep{Kot2021-xe}. Non-contact testing techniques are also considered a viable and integral part of SHM, though they present challenges in noise reduction and data processing \citep{Abu_Dabous2020-qp}. The rapid advancement of sensor technologies, such as vibration, acoustic, and thermal sensors, is central to this field. This progress also brings a focus on new research directions, including sensor development for harsh environments, integration with machine learning, and reducing power consumption \citep{Hassani2023-hi}.

Driving the field forward, recent research proposes innovative methods to overcome traditional limitations. One such proposal involves attaching piezoelectric sensors to unmanned aerial vehicles to detect structural defects without expensive image processing \citep{Na2016-vf}. Another novel method embeds piezoelectric sensors within concrete fiber content, using an ensemble machine learning model to classify signals and evaluate subsurface defects, which could be a valuable alternative for new construction \citep{Yang2022-nl}. The integration of ML continues to prove effective for specific material properties. For instance, \citet{Hola2023-sh} determined that random forest algorithms perform best in predicting the moisture content of concrete and masonry. As these novel techniques mature, \citet{Guemes2020-yl} stressed that establishing rigorous standards, such as the probability of detection, is essential for validating their performance and ensuring industry adoption.

The convergence of these advanced sensing technologies facilitates the transition to NDE 4.0. The integration of NDI within a Digital Twin ecosystem is a fundamental enabler of ALM \citep{Hassani2023-hi}, which seeks to optimize the value and safety of a physical asset from its initial manufacturing to its eventual decommissioning \citep{Meyendorf2023-qv}. By utilizing continuous or on-demand data from NDE 4.0 systems, ALM transitions from traditional, schedule-based maintenance to a predictive paradigm. For example, in the case of civil infrastructure like bridges, real-time sensor data can be used to update structural simulations \citep{Abu_Dabous2020-qp, Khedmatgozar_Dolati2021-as}, allowing engineers to predict the RUL of specific components rather than relying on reactive repairs after a defect is detected. This data-driven approach ensures maintenance resources are allocated precisely when needed, extending the operational lifespan of the asset while maintaining rigorous safety standards.

\section{Literature Type in NDI Research} \label{sec:review-literature-type-in-ndi-research}

The academic landscape of NDI research is defined not only by the technologies under investigation but also by the fundamental objectives that drive scientific inquiry. To gain a deeper understanding of the field's maturity and trajectory, we categorize the reviewed literature according to its primary contribution, including comprehensive reviews that consolidate existing knowledge, proposals for novel methodologies that push technical boundaries, and rigorous evaluations that benchmark performance. This structural classification reveals the dynamic interplay between theoretical innovation and practical validation, offering a perspective on how NDI research evolves from initial concept to standardized industrial practice. A detailed breakdown of this body of work is presented in Table \ref{tab:review-literature-type-paper-domain}, which maps each study to its specific research contribution.

\begin{longtable}[c]{lp{0.8cm}p{0.8cm}p{0.8cm}p{0.8cm}p{0.8cm}p{0.8cm}}
    \caption{Domains of the research papers discussed in Section \ref{sec:review-literature-type-in-ndi-research}. Domain labels are R = Review, PNM = Proposal of New Method, and EM = Evaluating Methods.\label{tab:review-literature-type-paper-domain}}\\
    \toprule
         Paper & R & PNM & EM \\
    \midrule
    \endfirsthead

    \toprule
         Paper & R & PNM & EM \\
    \midrule
    \endhead
         \citet{Lange1994-nf} & \cellcolor{cellbg} & & \\
         \citet{Sezgin2001-cc} & & & \cellcolor{cellbg} \\
         \citet{Huang2003-gy} & & \cellcolor{cellbg} & \\
         \citet{Ruzek2006-pq} & & & \cellcolor{cellbg} \\
         \citet{Farrar2007-jo} & \cellcolor{cellbg} & & \\
         \citet{Achenbach2009-jj} & \cellcolor{cellbg} & & \\
         \citet{Rummel2012-zl} & & & \cellcolor{cellbg} \\
         \citet{Lee2014-xt} & \cellcolor{cellbg} & & \\
         \citet{Usamentiaga2014-hh} & & & \cellcolor{cellbg} \\
         \citet{Du_Plessis2016-qn} & & & \cellcolor{cellbg} \\
         \citet{Gholizadeh2016-ww} & \cellcolor{cellbg} & & \\
         \citet{Na2016-vf} & & \cellcolor{cellbg} & \\
         \citet{Chauveau2018-qo} & \cellcolor{cellbg} & & \\
         \citet{Dwivedi2018-ts} & \cellcolor{cellbg} & & \\
         \citet{Akcay2019-xy} & & \cellcolor{cellbg} & \\
         \citet{Zaporozhets2019-cn} & & \cellcolor{cellbg} & \\
         \citet{Abu_Dabous2020-qp} & \cellcolor{cellbg} & & \\
         \citet{Al_Rahman_M_Abu_Ebayyeh2020-ox} & \cellcolor{cellbg} & & \\
         \citet{El_Masri2020-ux} & \cellcolor{cellbg} & & \cellcolor{cellbg} \\
         \citet{Baumgartl2020-fe} & & \cellcolor{cellbg} & \\
         \citet{Guemes2020-yl} & \cellcolor{cellbg} & & \\
         \citet{Honarvar2020-tw} & \cellcolor{cellbg} & & \\
         \citet{Mishra2020-qv} & & & \cellcolor{cellbg} \\
         \citet{Sreeshan2020-dc} & & \cellcolor{cellbg} & \\
         \citet{Wang2020-up} & \cellcolor{cellbg} & & \\
         \citet{Alloqmani2021-gz} & \cellcolor{cellbg} & & \\
         \citet{Deepak2021-pa} & & & \cellcolor{cellbg} \\
         \citet{Fang2021-nw} & & \cellcolor{cellbg} & \\
         \citet{Fernandes2021-sf} & & \cellcolor{cellbg} & \\
         \citet{Hashimoto2021-mf} & & \cellcolor{cellbg} & \\
         \citet{Khedmatgozar_Dolati2021-as} & \cellcolor{cellbg} & & \\
         \citet{Kot2021-xe} & & & \cellcolor{cellbg} \\
         \citet{Na2021-eq} & & \cellcolor{cellbg} & \\
         \citet{Song2021-ir} & & \cellcolor{cellbg} & \\
         \citet{Gupta2022-id} & \cellcolor{cellbg} & & \\
         \citet{Posilovic2022-ay} & \cellcolor{cellbg} & & \\
         \citet{Yang2022-nl} & & \cellcolor{cellbg} & \\
         \citet{Angermann2023-at} & & \cellcolor{cellbg} & \\
         \citet{Bajgholi2023-sr} & & & \cellcolor{cellbg} \\
         \citet{Hassani2023-hi} & \cellcolor{cellbg} & & \\
         \citet{Hola2023-sh} & & & \cellcolor{cellbg} \\
         \citet{Jodhani2023-mn} & \cellcolor{cellbg} & & \\
         \citet{Lin2023-fd} & \cellcolor{cellbg} & & \\
         \citet{Sun2023-yp} & \cellcolor{cellbg} & & \\
         \citet{Yang2025-rz} & & & \cellcolor{cellbg} \\
    \bottomrule
    
\end{longtable}

\subsection{Review} \label{sec:review-review}

Review articles are fundamental to NDI research, providing comprehensive overviews of advancements, challenges, and the state of the art. General reviews, for example, have discussed the broad advancements in NDI, noting the field's maturity and concluding that no single technique is suitable for all situations, thus highlighting the benefits of combining methods \citep{Gupta2022-id}. Other broad reviews have identified common NDI methods and their application areas, such as in aerospace and manufacturing, while also cataloging their capabilities and limitations \citep{Dwivedi2018-ts}.

A substantial portion of review literature is dedicated to SHM. Early work in this area reviewed specific techniques like mechanical impedance analysis, noting its advantage of dry point contact, which makes it suitable for almost any material \citep{Lange1994-nf}. More foundational reviews identified the steps of a comprehensive SHM system and underscored persistent challenges, including structural degradation, the lack of quality defect data, and proving the cost benefit to industry \citep{Farrar2007-jo}. Others have also discussed the potential of NDI in SHM, weighing the benefits of increased safety and reduced costs against implementation challenges like sensor cost and data transmission \citep{Achenbach2009-jj}.

Reviews focusing on specific SHM applications are also common. \citet{Lee2014-xt} reviewed the history of bridge inspection in the U.S., classifying NDI methods for concrete and steel and noting that few states fully leverage NDI's potential. \citet{Khedmatgozar_Dolati2021-as} focused specifically on NDI techniques for local damage detection on steel bridges. Similarly, \citet{El_Masri2020-ux} explored the use of different NDI techniques in evaluating building thermodynamics performance. Other reviews have examined SHM from a technological perspective. \citet{Guemes2020-yl} highlighted the benefits of SHM for safety and reliability, discussing the probability of detection as a standard for evaluation. \citet{Abu_Dabous2020-qp} reviewed non-contact testing techniques, underscoring their viability as part of SHM while identifying challenges in data processing and validation. \citet{Hassani2023-hi} provided a comprehensive review of the rapid advancement in sensor technologies for SHM, discussing future research directions like developing sensors for harsh environments and integrating ML.

Another significant group of reviews centers on NDI for specific materials. Several authors have provided overviews of methods for composite materials \citep{Gholizadeh2016-ww, Wang2020-up}. \citet{Gholizadeh2016-ww} discussed seven common NDI methods, emphasizing the importance of correct technique selection. \citet{Jodhani2023-mn} focused specifically on ultrasonic techniques for composites, concluding that the future of this field will heavily depend on automation and AI. Similarly, NDI for additive manufacturing has been reviewed, with \citet{Chauveau2018-qo} providing a comprehensive overview of techniques for welded and additively manufactured parts, while \citet{Honarvar2020-tw} focused on the challenges of ultrasonic testing for 3D printed components.

Reviews also cover NDI in other specific industries, such as automated optical inspection in electronics manufacturing \citep{Al_Rahman_M_Abu_Ebayyeh2020-ox} and the use of machine learning for non destructive food quality testing \citep{Lin2023-fd}. Finally, a growing number of reviews focus on the algorithms themselves. \citet{Alloqmani2021-gz} studied anomaly detection methods in general and medical fields, finding DL to be the most popular approach. Other reviews have focused on specific algorithmic applications, such as the use of ML for ultrasonic weld inspection \citep{Sun2023-yp} and the performance of different DL models for anomaly detection in ultrasonic scans \citep{Posilovic2022-ay}.

\subsection{Proposal of New Method} \label{sec:review-proposal-of-new-method}

A significant body of NDI research is dedicated to proposing novel methods, ranging from advanced data processing algorithms to innovative sensor applications and inspection workflows. In the realm of data processing, for instance, \citet{Huang2003-gy} presented methods for visualizing noisy, high-resolution volume data from industrial CT. They proposed a combination of region growing techniques and a 2D histogram interface to facilitate volumetric feature extraction and modeling of concealed structures. Other research has focused on applying new computational models to specific NDI challenges. \citet{Baumgartl2020-fe} proposed a CNN approach to identify defects in the additive manufacturing process using thermographic imaging, achieving high accuracy with low computational cost. Similarly, \citet{Hashimoto2021-mf} used a deep generative model to identify minute 0.15 mm wide cracks in magneto-optical images of steel specimens.

DL models have been proposed for highly specialized tasks, such as inspecting cylinder liners using a CNN and a GAN to predict surface texture depth with micrometer accuracy \citep{Angermann2023-at}, or for thermal image augmentation to enhance defect visibility using GAN \citep{Fang2021-nw}. A prominent example of algorithmic innovation is the semi-supervised GANomaly method by \citet{Akcay2019-xy}, which uses a conditional GAN to effectively detect anomalies with very little labeled data. Data processing has also been improved in other ways, as \citet{Zaporozhets2019-cn} proposed using the Hilbert transform to extract more effective features from impedance signals in composite structures.

Beyond algorithms, many new methods involve novel sensor applications and inspection strategies. \citet{Yang2022-nl} proposed embedding a piezoelectric sensor within concrete fiber content, using an ensemble ML model to evaluate subsurface defects. \citet{Na2021-eq} also used an impedance-based technique, combining it with a probabilistic neural network to identify bolt loosening with high accuracy from a very small dataset. New methods have also been developed to solve specific field problems, like the electromagnetic induction impedance method proposed by \citet{Song2021-ir} to reduce noise when inspecting power station grounding grids. Innovative deployment strategies have also been suggested, such as attaching piezoelectric sensors to an unmanned aerial vehicle to detect structural defects without expensive image processing \citep{Na2016-vf}.

Finally, some research proposes new end-to-end inspection processes. \citet{Sreeshan2020-dc} presented a complete method for detecting defects in aerospace composites using pulsed thermography, involving preprocessing, defect identification, and segmentation to successfully find most simulated defects. Others have demonstrated the power of combining multiple NDI techniques. \citet{Fernandes2021-sf}, for example, presented a method to characterize ancient marquetry by integrating air-coupled ultrasound, X-ray microtomography, and infrared thermography, which proved to be a highly useful tool for identifying a wide range of defects like delaminations, cracks, and voids.

\subsection{Evaluating Methods} \label{sec:review-evaluating-methods}

A critical component of NDI research involves the formal evaluation of existing methods to compare their performance and identify their specific strengths and weaknesses. This evaluation often takes the form of direct comparisons for a specific task. For example, \citet{Ruzek2006-pq} tested visual inspection, ultrasonic C-scan, and shearography to identify impact damage, concluding that shearography performed best in their case. Similarly, \citet{El_Masri2020-ux} evaluated six NDI techniques, namely, Ground Penetrating Radar (GPR), Light Detection and Ranging (LiDAR), Ultrasound, Close-Range Photogrammetry, and Through-Wall Imaging Radar (TWIR). \citet{Deepak2021-pa} also evaluated five different NDI techniques for inspecting low-carbon steel weld joints and determined that the best method depends entirely on the specific application requirements.

Other studies focus on a deep evaluation of a single technology or its application. \citet{Usamentiaga2014-hh} reviewed the use of infrared thermography, noting its advantages as a contactless method for hazardous areas but also highlighting its limitations, such as vulnerability to environmental factors like ambient temperature and radiation, and a dependency on material surface emissivity. \citet{Kot2021-xe} reviewed a suite of NDI techniques for SHM, discussing challenges like data interoperability and cost to evaluate their potential for infrastructure maintenance. In a unique comparison, \citet{Du_Plessis2016-qn} assessed the performance of medical versus industrial CT scanners. They proposed that medical scanners, despite lower resolution, could be a viable and cost-effective alternative for large objects where high throughput is needed. A comparison of NDI techniques on weld joints in hydraulic turbine runners by \citet{Bajgholi2023-sr} compared conventional UT, radiography testing, PAUT, and PAUT using TFM. They found PAUT using TFM outperforms other methods in detecting flaws of $3 mm$ or smaller.

As the field becomes increasingly digitized, the evaluation of computational algorithms has emerged as a key research area. In the domain of ML, classical algorithms have demonstrated remarkable resilience against more complex architectures. \citet{Hola2023-sh} found that the random forest algorithm performed best in predicting the moisture content of concrete and masonry. This finding was corroborated by \citet{Yang2025-rz}, who rigorously evaluated several ML algorithms across two datasets and observed that Random Forest consistently outperformed more complex models.  Other studies have focused on specific predictive tasks. \citet{Sezgin2001-cc} conducted an extensive survey of image thresholding methods for NDI, using performance measures like misclassification error to identify a robust and generalizable mixture method. In the ML domain, \citet{Mishra2020-qv} evaluated three models for predicting the compressive strength of historic masonry. They determined that a support vector regression model offered a more practical and accurate solution than both artificial neural networks and adaptive neuro fuzzy inference systems for their application.

Finally, some literature focuses on standardizing the evaluation process itself. \citet{Rummel2012-zl}, for example, discussed the development and application of the Probability of Detection (POD) metric, outlining its history and highlighting its importance as an ongoing standard for NDI reliability.

\section{Discussion and Concluding Remarks} \label{sec:review-discussion}
This study provides a comprehensive review of the non-destructive inspection (NDI) field by synthesizing a wide range of academic research. We organized our analysis across three primary axes, which are the computational algorithms that automate inspection, the physical technologies deployed for data acquisition, and the key industrial sectors where NDI is an essential process. Furthermore, we categorized the reviewed literature to understand its focus by distinguishing between foundational reviews, proposals of new methods, and evaluations of existing techniques.

A key finding of this review is the increasing prominence of advanced algorithms in NDI, particularly deep learning models. As detailed in Table \ref{tab:review-algorithm-pro-con}, these models have become favored for their potential to deliver high accuracy and superior performance in complex defect detection tasks. However, this trend is accompanied by significant challenges associated with the transition to NDE 4.0. The reliability of most machine learning and deep learning algorithms is heavily dependent on the availability of large, high-quality, and well-annotated training datasets. This data requirement remains a primary barrier that limits their widespread and dependable implementation in diverse real-world scenarios. Furthermore, the shift toward NDE 4.0 introduces complexities regarding data interoperability and cybersecurity as legacy systems must be integrated into modern digital twin ecosystems. The industry faces a critical need to standardize digital protocols to ensure that automated inspection data can be securely and effectively utilized across the manufacturing lifecycle.

\begin{table}[pos=htbp]
    \centering
    \caption{Benefits and possible improvements for different algorithms in NDI.}
    \label{tab:review-algorithm-pro-con}
    \begin{tabular}{p{2.5cm}p{4cm}p{4cm}}
    \toprule
        Algorithm & Benefits & Possible improvements \\
    \midrule
        Computer Vision and Image Processing & {
            - Automation of visual inspection \cite{Brosnan2004-yd, Sreeshan2019-sf, Mahanti2022-sp}.\newline
            - Better speed, consistency, and objectivity compared to human inspector \cite{Brosnan2004-yd, Mahanti2022-sp}.\newline
            - Understandability and interpretability for human inspectors \cite{Huang2003-gy, Sreeshan2019-sf}.
        } & {
            - The algorithms depend on proper illumination and other image acquisition environments \cite{Brosnan2004-yd}.\newline
            - Noise reduction in images from processed signals \cite{Huang2003-gy}.\newline
            - Computational cost is quite high \cite{Yousefi2018-el}.\newline
            - No general model for all different defects \cite{Mahanti2022-sp}.\newline
            - High cost of equipment capable of handling real-world noise \cite{Memmolo2015-er}.
        } \\
     \midrule
        Machine Learning & {
            - Ability to make predictions based on limited data \cite{Mishra2020-qv, Gardner2020-bn}.\newline
            - Processing large amount of data, localization of damage, automated damage detection \cite{Gardner2020-bn, Taheri2022-ix}.\newline
            - Increased accuracy compared to other automated techniques \cite{Taheri2022-ix}.
        } & {
            - Lack of quality training data \cite{Gardner2020-bn, Taheri2022-ix}.\newline
            - Models depend on pre-processing the data before training \cite{Taheri2022-ix}.\newline
            - There is a high computational cost for training with a large amount of data \cite{Taheri2022-ix}.\newline
            - Interpretability of the models is low \cite{Lin2023-fd}.\newline
            - Model accuracy depends on appropriate data selection for training \cite{Sun2023-yp}.
        } \\
    \midrule
        Deep Learning & {
            - Higher accuracy and performance compared to traditional models \cite{Yousefi2018-el, Akcay2019-xy, Ma2020-cu, Siegel2020-cp, Presenti2022-kp, Lin2023-fd}.\newline
            - Are capable of detecting the severity of the defects \cite{Siegel2020-cp}.\newline
            - Noise tolerant \cite{Presenti2022-kp}.
        } & {
            - Requires more training data \cite{Lin2023-fd}.\newline
            - Requires higher computational resources and training time \cite{Lin2023-fd}.\newline
            - Model accuracy depends on appropriate hyperparameter tuning \cite{Sun2023-yp}.
        }\\
    \midrule
        Anomaly Detection & {
            - High accuracy with a limited amount of labeled data \cite{Akcay2019-xy, Baumgartl2020-fe}.\newline
            - Low computational cost \cite{Baumgartl2020-fe}.\newline
            - Noise tolerant \cite{Presenti2022-kp}.
        } & {
            - Lack of labeled data and benchmark \cite{Alloqmani2021-gz}.
        }\\
    \midrule
        Generative Model & {
            - Synthetic data addressing the scarcity of training data \cite{Akcay2019-xy}.
        } & {
            - Performance is highly dependent on the generated loss function and can be difficult to generalize for a large dataset \citep{Fang2021-nw}.
        } \\
    \bottomrule
    \end{tabular}
\end{table}

Beyond the computational methods, our analysis of physical NDI technologies reveals distinct adoption patterns. Ultrasound-based techniques remain among the most popular, striking an effective balance between application performance and operational cost. Visual inspection continues to be a widely used, cost-effective approach for surface-level assessments, though its inherent inability to detect subsurface defects is a major limitation. A detailed comparison of the benefits and limitations for these and other core technologies is presented in Table \ref{tab:review-technology-pro-con}.

\begin{table}[pos=htbp]
    \centering
    \caption{Benefits and limitations of different technologies in NDI.}
    \label{tab:review-technology-pro-con}
    \begin{tabular}{p{2.5cm}p{4cm}p{4cm}}
    \toprule
        Technology & Benefits & Limitations \\
    \midrule
        Ultrasound & {
            - Ability to detect subsurface defects \cite{Dwivedi2018-ts}.\newline
            - Works on a variety of materials and arbitrary shapes in great detail \cite{Wang2008-hg, El_Masri2020-ux}.\newline
            - Internal access is not necessary \cite{Wang2020-up}.
        } & {
            - Requires the material to be sound conductive \cite{Dwivedi2018-ts}.\newline
            - Possible to miss micro defects \cite{Nsengiyumva2021-bo, Gupta2022-id}.\newline
            - Requires processing and analyzing signals and depends on the expertise of the inspector \cite{Gupta2022-id}.\newline
            - Sensitive to environment \cite{Wang2020-up}.
        } \\
    \midrule
        Radiography and X-Ray & {
            - Ability to detect subsurface defects \cite{Wang2020-up, Gupta2022-id}.\newline
            - Works on variety of materials \cite{Wang2020-up, Nsengiyumva2021-bo, Gupta2022-id}.
        } & {
            - Accuracy drops for complex structures \cite{Wang2020-up}.\newline
            - Expensive technology \cite{Gupta2022-id}.\newline
            - The process is slow \cite{Nsengiyumva2021-bo}.\newline
            - Access to both sides of the material is required \cite{Wang2020-up, Nsengiyumva2021-bo, Gupta2022-id}.\newline
            - Safety hazard for human inspectors \cite{Wang2020-up, Nsengiyumva2021-bo, Gupta2022-id}.
        } \\
    \midrule
        Infrared, Thermal, and Low Frequency & {
            - Dry surface contact is adequate for low-frequency NDI techniques \cite{Cawley1984-le}.\newline
            - Internal access is not necessary \cite{Wang2020-up}.\newline
            - Cost effective \cite{Wang2020-up}.
        } & {
            - Defect shape and depth are not precise \cite{Wang2020-up, Nsengiyumva2021-bo}.
        } \\
    \midrule
        Visual Inspection & {
            - Simple, fast, and low cost \cite{Wang2020-up, Nsengiyumva2021-bo}.\newline
            - Non-contact testing \cite{Nsengiyumva2021-bo}.
        } & {
            - Usually works only for surface defect detections \cite{Wang2020-up, Nsengiyumva2021-bo, Gupta2022-id}.\newline
            - Micro-defects are usually ignored \cite{Wang2020-up}.\newline
            - Detection is not very reliable and subjective \cite{Ruzek2006-pq, Wang2020-up}.
        } \\
    \midrule
        Electromagnetic Field & {
            - Usually, dry-contact testing, making it a useful technique for many conductive materials \cite{Lange1994-nf}.\newline
            - Detects subsurface defect location and shape in ferromagnetic materials \cite{Kosmas2007-uf}.
        } & {
            - Requires conductive material \cite{Nsengiyumva2021-bo}.\newline
            - Sensitive to electromagnetic interference \cite{Song2021-ir}.
        } \\
    \bottomrule
    \end{tabular}
\end{table}

The practical impact of these algorithms and technologies becomes most clear when examining their industrial applications. NDI has solidified its role as an integral component in sectors ranging from structural health monitoring and aerospace to food safety and the conservation of fragile artifacts. As outlined in Table \ref{tab:review-application-pro-con}, each application area presents a unique set of benefits and operational challenges. While many such obstacles must still be overcome, the undeniable advantages of NDI in ensuring safety, quality, and reliability have confirmed its necessity.

\begin{table}[pos=htbp]
    \centering
    \caption{Benefits and challenges of application areas in NDI.}
    \label{tab:review-application-pro-con}
    \begin{tabular}{p{2.5cm}p{4cm}p{4cm}}
    \toprule
        Application areas & Benefits & Challenges \\
    \midrule
        Composite Materials & {
            - Detecting small surface-level and subsurface defects, cracks, porosity, delamination, and others requires NDI \cite{Diamanti2005-mr, Gholizadeh2016-ww, Siegel2020-cp}.\newline
            - Different NDI techniques are available to identify defects in materials with complex shapes \cite{Honarvar2020-tw}.
        } & {
            - Most NDI techniques require additional processing of signals and data \cite{Honarvar2020-tw}.\newline
            - Requires balancing between cost and efficiency \cite{Diamanti2005-mr, Gholizadeh2016-ww}.
        }\\
    \midrule
        Organic Material and Food Safety & {
            - Ensures speed, consistency, and objectivity of inspections \cite{Brosnan2004-yd}.\newline
            - Real-time damage detection, including internal defects \cite{Mahanti2022-sp}.
        } & {
            - Quality data to train automated models and interoperability is a challenge \cite{Lin2023-fd}.
        }\\
    \midrule
        Construction Material & {
            - Different NDI methods offer on-site inspection where access to only one side is sufficient \cite{Ould_Naffa2002-ab}.
        } & {
            - Varying penetration depth and difference in materials make it challenging to have a universal inspection method \cite{Ould_Naffa2002-ab, Deepak2021-pa}.
        }\\
    \midrule
        Fragile Artifacts & {
            - NDI is essential to evaluate the strength, weathering, and presence of defects in fragile artifacts without damaging or degrading the artistic values \cite{Martinho2014-kh, Mishra2020-qv}.
        } & {
            - On-site inspections are vulnerable to the environment and require special equipment, which can be expensive \cite{Memmolo2015-er}.
        }\\
    \midrule
        Medical & {
            - Higher throughput of medical scanning equipment can be useful where high resolution is not needed \cite{Du_Plessis2016-qn}.
        } & {
            - Low penetration depth due to low voltage \cite{Du_Plessis2016-qn}.
        }\\
    \midrule
        Structural Health Monitoring & {
            - Safety and structural integrity of buildings after natural events like earthquakes, or routine check of infrastructure integrity requires NDI techniques \cite{Farrar2007-jo}.
        } & {
            - Data interoperability, penetration depth, and cost are major challenges \cite{Kot2021-xe}.\newline
            - Research in modern eco-friendly sensors adaptive to the environment is necessary \cite{Hassani2023-hi}.
        }\\
    \bottomrule
    \end{tabular}
\end{table}

\subsection{Research Gaps and Future Directions}\label{sec:review-gaps-and-future}
Despite the significant advancements highlighted in this review, the transition to NDE 4.0 has exposed several critical gaps in the current body of knowledge that must be addressed to fully realize the potential of the automated NDI process.

\textbf{Data Scarcity and Standardization:} A primary deficiency is the chronic lack of standardized, open-access datasets for benchmarking novel algorithms. Unlike the computer vision community, which benefits from massive public repositories like ImageNet, the NDI community often relies on proprietary, small-scale, or synthetic datasets. This fragmentation makes it nearly impossible to rigorously compare the generalizability of different DL models. Future research must prioritize the creation of shared, anonymized repositories of defect data to accelerate algorithmic development. Furthermore, research into few-shot learning and transfer learning techniques specifically tailored for NDI spectral data is urgently needed to mitigate the dependency on large labeled datasets.

\textbf{Interpretability and Trust (XAI):} As DL models achieve exceptional accuracy, their ``black box" nature becomes a liability, particularly in safety-critical industries like aerospace and nuclear power. Current research often prioritizes classification accuracy over interpretability. There is a pressing need for Explainable AI (XAI) methodologies that can visualize the decision-making process of a model, mapping specific signal features to physical defect characteristics. Developing verifiable and interpretable models is a prerequisite for regulatory certification and industrial trust.

\begin{table}[pos=htbp]
    \centering
    \caption{Research gaps and future research opportunities in NDI.}
    \label{tab:review-research-gaps}
    \begin{tabular}{p{2.5cm}p{4cm}p{4cm}}
    \toprule
        \textbf{Research Area} & \textbf{Identified Gap / Limitation} & \textbf{Future Research Idea} \\
    \midrule
        \textbf{Deep Learning Algorithms} & 
        Models act as ``black boxes,'' lacking transparency in how defect classifications are derived, which hinders adoption in safety-critical sectors. & 
        Development of Explainable AI (XAI) frameworks specifically for NDI that provide visual or feature-based justification for defect classification. \\
    \midrule
        \textbf{Data Availability} & 
        Severe scarcity of labeled industrial defect data, and reliance on synthetic data that may not perfectly mimic real-world physics. & 
        Investigation into physics-informed neural networks (PINNs) and federated learning to train robust models without sharing proprietary data. \\
    \midrule
        \textbf{Sensing Techniques} & 
        Single-modality inspections have inherent physical blind spots regarding depth or material type. & 
        Development of real-time multi-modal sensor fusion algorithms that combine surface and subsurface data. \\
    \midrule
        \textbf{Structural Health Monitoring (SHM)} & 
        Sensors in harsh environments suffer from power limitations and degradation over time, and the data transmission bandwidth is limited. & 
        Research into self-powered, eco-friendly piezoelectric sensors and edge-computing algorithms that process data on-chip to reduce transmission load. \\
    \midrule
        \textbf{NDE 4.0 Ecosystem} & 
        Lack of standardized protocols for integrating NDI data into digital twins, restricting the data siloed in inspection reports. & 
        Creation of universal protocols that allow automated NDI results to dynamically update the structural simulation of a digital twin. \\
    \bottomrule
    \end{tabular}
\end{table}

\textbf{Sensor Fusion and Multi-Modality:} While individual NDI techniques are mature, single-sensor inspections often suffer from blind spots. For example, visual inspection misses subsurface voids, while ultrasound struggles with complex surface geometries. Current research rarely explores the automated, real-time fusion of disparate sensor data, such as combining thermographic video with ultrasonic waveforms. Future work should focus on multi-modal learning architectures that can spatially register and fuse data from complementary sensors to provide a holistic assessment of structural integrity.

\textbf{Integration with Digital Twins:} The concept of NDE 4.0 relies on the seamless integration of inspection data with Digital Twins. However, current literature highlights a significant gap in data interoperability and standard protocols. Inspection data remains siloed, disconnected from the life-cycle management of the inspected object. Research is needed to develop standardized digital threads that allow NDI data to automatically update a digital twin, enabling real-time prognosis of remaining useful life (RUL) rather than simple defect detection.

A structured summary of these identified gaps and potential research avenues is presented in Table \ref{tab:review-research-gaps}.

\begin{figure}[pos=htbp]
    \centering
    \includegraphics[width=.85\linewidth]{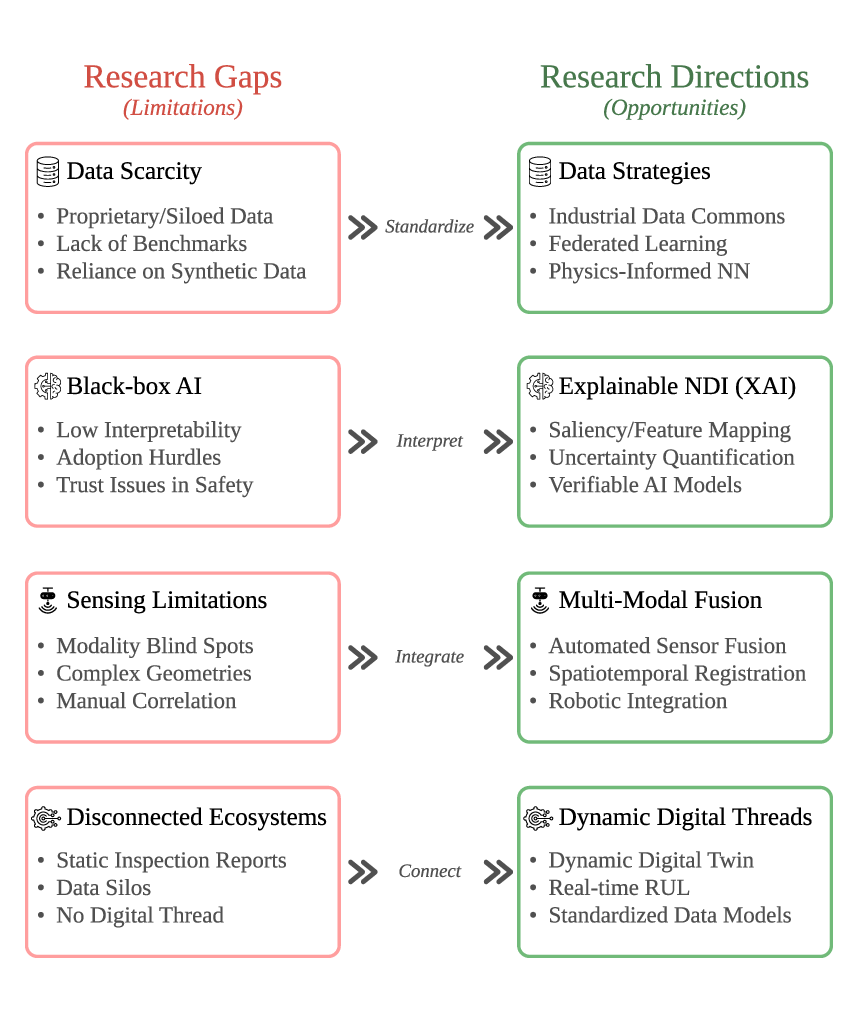}
    \caption{Strategic research roadmap illustrating the transition from current limitations to NDE 4.0 opportunities.}
    \label{fig:review-roadmap}
\end{figure}

Complementing the itemized challenges in Table \ref{tab:review-research-gaps}, Figure \ref{fig:review-roadmap} provides a visual synthesis of the strategic transition toward NDE 4.0. This diagram illustrates how currently fragmented input domains, such as \textit{data}, \textit{algorithms}, \textit{sensing}, and \textit{ecosystems}, must structurally converge through targeted research strategies. By mapping specific limitations to their corresponding opportunities, the figure highlights that resolving these gaps is the way to unlock a fully integrated, autonomous, and digital-twin-enabled future.

\subsection{Concluding Remarks} \label{sec:review-conclusion}

This review has charted the current landscape of non-destructive inspection, revealing a field in the midst of a radical paradigm shift. We have moved beyond the era where NDI was merely a support process for quality control. It is now a primary driver of the Industry 4.0 revolution. The analysis of algorithmic, technological, and industrial perspectives confirms that the boundaries between physical inspection and digital modeling are irrevocably blurring.

The most significant bottleneck in modern NDI is no longer the sensitivity of the sensors, but the intelligence of the interpretation. While traditional techniques like ultrasound and radiography remain physically robust, their future utility depends entirely on their integration with intelligent algorithms capable of navigating the noise and complexity of real-world environments. The challenges identified in this study, from the data dependency of deep learning to the need for standardized digital protocols, should not be viewed as roadblocks but as the defining frontiers of the next generation of research. Ultimately, the future of NDI lies in augmenting flaw detection with autonomous, data-driven assurance of structural integrity, leading to a safer and more efficient world.

\section*{Statements \& Declarations}\label{review-declaration}
\subsection*{Funding}
This work was supported in part by the U.S. Department of Energy, Office of Science, under contract \textit{DE-AC02-06CH11357}.

\subsection*{Competing Interests}
The authors have no relevant financial or non-financial interests to disclose.




\end{document}